%% file: main.tex
\pdfoutput=1

\documentclass[10pt,twocolumn,letterpaper]{article}

\usepackage[pagenumbers]{cvpr}

\input{preamble}

\definecolor{cvprblue}{rgb}{0.21,0.49,0.74}
\usepackage[pagebackref,breaklinks,colorlinks,allcolors=cvprblue]{hyperref}
\renewcommand{\appref}[1]{\hyperref[app:#1]{Appendix~#1}}
\hypersetup{pdftitle={Say Less, Hallucinate Less? Diagnosing and Restoring Object Coverage in Preference-Optimized Vision-Language Models}, pdfauthor={Hansang Lee}}

\title{Say Less, Hallucinate Less? Diagnosing and Restoring Object Coverage\\ in Preference-Optimized Vision-Language Models}

\author{Hansang Lee\\
Department of Computer Science, Seoul Women's University, Seoul, Republic of Korea\\
{\tt\small hansanglee@swu.ac.kr}
}

\begin{document}
\maketitle
\input{sec/0_abstract}
\input{sec/1_intro}
\input{sec/2_method}
\input{sec/3_setup}
\input{sec/4_gain}
\input{sec/5_recall}
\input{sec/6_restore}
\input{sec/7_discussion}
\FloatBarrier
\input{sec/8_llm}
{
    \small
    \bibliographystyle{ieeenat_fullname}
    \bibliography{main}
}

\appendix
\crefname{section}{Appendix}{Appendices}
\Crefname{section}{Appendix}{Appendices}
\crefname{appendix}{Appendix}{Appendices}
\Crefname{appendix}{Appendix}{Appendices}
\renewcommand{\thesection}{\Alph{section}}
\renewcommand{\thetable}{\Alph{section}\arabic{table}}
\renewcommand{\thefigure}{\Alph{section}\arabic{figure}}
\makeatletter
\@addtoreset{table}{section}
\@addtoreset{figure}{section}
\makeatother
\renewcommand{\dblfloatpagefraction}{0.7}
\renewcommand{\floatpagefraction}{0.7}
\section*{Appendix}
\pdfbookmark[0]{Appendix}{appendix}

The appendices follow the letters used in the main text.
\Cref{app:A,app:B} give the checks of the implementation and of the post-hoc EOS cut, \cref{app:C,app:D,app:E,app:F,app:G,app:H} the auxiliary analyses of the gain, \cref{app:I} the negative EOS probe, \cref{app:J} the batch-size check, \cref{app:K} the details and controls of restoration, and \cref{app:L} the held-out decomposition.
Unless stated otherwise, intervals are nominal 95 percent image bootstrap intervals with 1,000 draws.

\input{sec/S1_appendix_A_F}

\input{sec/S2_appendix_G_L}

\end{document}

%% file: preamble.tex
\usepackage{cuted} %

\usepackage{currfile} %

\usepackage{caption} %

\usepackage{amsmath,amssymb}
\usepackage{booktabs}
\usepackage{array}
\usepackage{multirow}
\usepackage{xspace}
\usepackage{needspace}
\newcommand{\Hu}{H_{\mathrm{u}}}
\newcommand{\Draw}{\Delta_{\mathrm{raw}}}
\newcommand{\DS}{\Delta_{S}}
\newcommand{\DR}{\Delta_{R}}

\newcommand{\appref}[1]{Appendix~#1}
\usepackage{placeins}
\usepackage{tikz}
\usetikzlibrary{positioning,arrows.meta}

%% file: sec/0_abstract.tex
\begin{abstract}
Preference optimization (PO) is widely used to reduce object hallucination in large vision-language models. We study four public PO checkpoints of LLaVA-1.5-7B. They name fewer object categories than the base model, and their lower hallucination rates bring no significant gain in set-level object F1 on CHAIR. We therefore ask how much of the PO gain the base model can reproduce by stopping earlier. A positive end-of-sequence (EOS) logit bias brings the base model to the recall of each PO model, and this early stopping alone reproduces 58 to 86 percent of the CHAIRi gain of three checkpoints on two disjoint COCO sets. For these three checkpoints on CHAIR-500, matching the caption length with the same EOS bias reproduces only 0 to 22 percent. In the reverse direction, a negative EOS bias under greedy decoding recovers only 2 to 9 percent of the lost CHAIR recall of these checkpoints, and 94 to 96 percent of the added object mentions repeat categories that the caption already names. The base caption of the same image still names about half of the missed CHAIR categories. We therefore propose base-guided object-category restoration, which appends high-confidence base-only categories to the PO caption without training. With thresholds fixed on a development set, it brings three checkpoints within one point of the base recall on a held-out COCO set with a lower distinct-object hallucination rate than the base model. At matched recall, its hallucination rate is also lower than the base stopping curve for RLAIF-V near the base recall and for RLAIF-V and CHAIR-DPO at an intermediate recall. We will release the code of the recall-controlled evaluation.
\end{abstract}

%% file: sec/1_intro.tex
\section{Introduction}
\label{sec:intro}

\begin{figure*}[t]
  \centering
  \includegraphics[width=\linewidth]{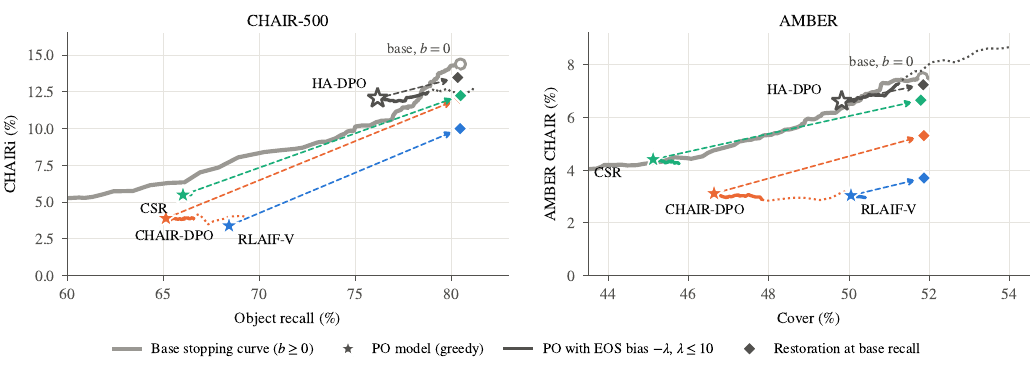}
  \caption{Base stopping reproduces much of the CHAIRi gain of RLAIF-V, CHAIR-DPO and CSR, and a negative EOS bias recovers little of their lost coverage. Hallucination against coverage for LLaVA-1.5-7B and four PO checkpoints on the development sets of CHAIR-500 (left) and AMBER (right). The gray curve is the base model with a positive EOS logit bias, and the vertical gap between a star and the curve at the same recall is the stopping-controlled gain. The short colored paths show each PO model with a negative EOS bias of up to 10, dotted where more than 5 percent of the captions reach the 512-token limit. The diamonds show cross-fitted restoration of each PO model aimed at the base recall, and the dashed arrows show the part of the PO gain that restoration gives up to recover the lost recall.}
  \label{fig:teaser}
\end{figure*}

Large vision-language models (LVLMs) often mention objects that are not in the image.
This object hallucination is measured on captions with CHAIR~\cite{rohrbach2018chair} and AMBER~\cite{wang2023amber}, and on yes/no questions with POPE~\cite{li2023pope}.
Preference optimization (PO) is widely used to reduce it.
Recent PO methods build their preference data from human or AI feedback~\cite{yu2024rlhfv,yu2025rlaifv}, synthetic dispreferred responses~\cite{zhou2024povid,zhao2023hadpo}, self-rewarding~\cite{zhou2024csr} or a CHAIR-based reward~\cite{compagnoni2025chairdpo}, and they share objectives in the style of direct preference optimization (DPO)~\cite{rafailov2023dpo}.
These methods and training-free decoding methods such as VCD~\cite{leng2024vcd} and GACD~\cite{wang2026gacd} report their gains on the same metrics.
A hallucination rate counts the wrong objects among the mentioned objects, so a model that mentions fewer objects can lower this rate without seeing the image better.

Recent studies point to this coverage side of hallucination.
Fazli \etal~\cite{fazli2026safe} report that the gains of recent inference-time mitigation methods come together with less informative outputs, and Sandhu \etal~\cite{sandhu2026heads} find shorter captions with lower object recall after an attention-head intervention on LLaVA.
For trained models, CHiP~\cite{fu2025chip} reports a lower AMBER Cover at a similar output length on LLaVA-1.6, Long \etal~\cite{long2026saying} study captions that say less after on-policy reinforcement learning on Qwen2.5-VL, and concurrent work by Alqahtani \etal~\cite{alqahtani2026beyond} reports a lower AMBER Cover after DPO training of LLaVA-1.5-7B.
Yue \etal~\cite{yue2024less} show that a model trained to stop earlier hallucinates less, and hallucinated objects appear more often late in a caption~\cite{zhou2024lure,zheng2025haltrapper,han2026egvca}.
Longer descriptions also contain more hallucinated objects, so LeHaCE~\cite{wei2024lehace} compares LVLMs at the same description length, as length-controlled evaluation does in LLM alignment~\cite{dubois2024lcalpaca}.
These studies report the coverage loss, train the stopping decision or control the caption length.
Concurrent work by Bu and Gao~\cite{bu2026buy} compares preference-trained LLaVA-1.5-7B models with base captions truncated to the same object recall and asks how much of the remaining reduction of hallucinated objects a small frozen probe reproduces.
We study two linked questions with one decoding control, the EOS logit bias.
How much of the hallucination gain of trained PO checkpoints can a base model reproduce by stopping earlier? Does their lost recall come back when they stop later?

We study four public PO checkpoints of LLaVA-1.5-7B~\cite{liu2024llava15}, namely RLAIF-V, CHAIR-DPO, CSR and HA-DPO.
All four lower CHAIRi and object recall together, and the recall drops from 80.5 to between 65.1 and 76.2 on CHAIR-500 (\cref{tab:standard}).
Their captions stay long, but they name fewer object categories, and the usual summaries then disagree.
The distinct-object hallucination rate favors every PO model with an interval that excludes zero, while the set-level object F1 on CHAIR detects no significant gain for any of them and drops for CSR on a held-out COCO set.
For the first question, we let the base model stop earlier with a single positive end-of-sequence (EOS) logit bias until it reaches the recall of each PO model.
On two disjoint sets of 500 COCO images, this base stopping curve reproduces 58 to 86 percent of the CHAIRi gain of RLAIF-V, CHAIR-DPO and CSR without any training (\cref{fig:teaser,tab:share}).
On the development sets, matching the caption length with the same EOS bias reproduces only 0 to 22 percent, since the PO captions stay long. At equal recall they are 1.3 to 1.8 times longer than the captions on the curve, so a length-controlled comparison misses most of this effect.

For the second question, we make RLAIF-V, CHAIR-DPO and CSR write longer with a negative EOS bias of up to 10.
They recover only 2 to 9 percent of the lost CHAIR recall, and 94 to 96 percent of the added object mentions refer to object categories that the caption already names (\cref{tab:debias}).
Earlier stopping of the base model thus reproduces much of the PO gain, while extending the PO captions recovers little of their lost object recall (\cref{fig:pipeline}). The base caption of the same image still names about half of the COCO categories that these models miss on CHAIR.

We turn these observations into an evaluation and a training-free intervention.
Under greedy decoding, one pass of the base model gives its EOS stopping curve for every nonnegative bias.
Base-guided object-category restoration appends to the PO caption the base-only object categories that the base model generates with high confidence.
It keeps only part of the PO gain. At matched recall it stays below the base stopping curve for RLAIF-V near the base recall and for RLAIF-V and CHAIR-DPO at an intermediate recall, and we find no such advantage for CSR.

\begin{figure*}[t]
  \centering
  \resizebox{\linewidth}{!}{\input{fig/pipeline}}
  \caption{Stopping and base-guided restoration. (a) A positive EOS bias makes the base model stop earlier. Teal and red mark grounded and hallucinated objects for illustration only, and the EOS bias does not use the ground truth. (b) A negative EOS bias makes the PO caption of \cref{fig:qual}b longer. The charts show changes in points on CHAIR-500 (\cref{tab:share,tab:debias}), and the early stop in (a) is the base stopping curve at the recall of each PO model. (c) Restoration of the RLAIF-V caption of \cref{fig:qual}a. Blue and orange mark PO and base-only categories. Confidence is the base log-probability of the first token of the first mention, and $\tau = -1.29$ is fixed on the development set.}
  \label{fig:pipeline}
\end{figure*}

Our contributions are as follows.
\begin{itemize}
  \item We propose recall-controlled evaluation against the EOS stopping curve of the base model. On two disjoint COCO sets, this curve reproduces 58 to 86 percent of the CHAIRi gain of RLAIF-V, CHAIR-DPO and CSR, while matching the caption length reproduces only 0 to 22 percent on the development sets (\cref{sec:method,sec:gain}).
  \item We show that, under greedy decoding, a later stop with a negative EOS bias recovers only 2 to 9 percent of the lost CHAIR recall of RLAIF-V, CHAIR-DPO and CSR, while the base caption of the same image still names about half of the missed categories on CHAIR (\cref{sec:recall}).
  \item We propose base-guided object-category restoration, a minimal training-free intervention that fills these misses from the base caption of the same image. With its threshold and acceptance criterion fixed on a development set, it brings the three PO models within one point of the base recall on a held-out set with a lower distinct-object hallucination rate than the unbiased base model, although only 41 to 45 percent of the appended categories are correct (\cref{sec:restore}).
\end{itemize}

%% file: fig/pipeline.tex
\begin{tikzpicture}[
  font=\scriptsize,
  w/.style={draw=black!25, fill=black!4, minimum width=0.42cm, minimum height=0.36cm, inner sep=0pt},
  wf/.style={draw=black!12, fill=black!2, minimum width=0.42cm, minimum height=0.36cm, inner sep=0pt},
  ok/.style={draw=teal!70!black, fill=teal!18, rounded corners=1.5pt, minimum height=0.36cm, inner sep=1.5pt},
  bad/.style={draw=red!70!black, fill=red!14, rounded corners=1.5pt, minimum height=0.36cm, inner sep=1.5pt},
  okf/.style={draw=teal!30, fill=teal!5, text=black!40, rounded corners=1.5pt, minimum height=0.36cm, inner sep=1.5pt},
  badf/.style={draw=red!30, fill=red!5, text=black!40, rounded corners=1.5pt, minimum height=0.36cm, inner sep=1.5pt},
  po/.style={draw=blue!60!black, fill=blue!12, rounded corners=1.5pt, minimum height=0.36cm, inner sep=1.5pt},
  rep/.style={draw=blue!45, fill=blue!4, text=black!55, rounded corners=1.5pt, minimum height=0.36cm, inner sep=1.5pt, dashed},
  bo/.style={draw=orange!80!black, fill=orange!18, rounded corners=1.5pt, minimum height=0.36cm, inner sep=1.5pt},
  bof/.style={draw=black!25, fill=black!4, text=black!40, rounded corners=1.5pt, minimum height=0.36cm, inner sep=1.5pt},
  eos/.style={draw=black, fill=black, text=white, rounded corners=1pt, minimum height=0.36cm, inner sep=1.5pt, font=\tiny\bfseries},
  ttl/.style={font=\footnotesize, anchor=west},
  lab/.style={font=\scriptsize, anchor=east, text=black!70},
  num/.style={font=\footnotesize, anchor=west},
  sub/.style={font=\scriptsize, anchor=north west, text=black!65},
  box/.style={draw=black!35, fill=black!2, rounded corners=3pt},
  btl/.style={font=\scriptsize, anchor=north west, text=black!80},
  arr/.style={-{Stealth[length=4pt]}, thick, black!55},
  node distance=0.06cm
]
\path[use as bounding box] (-0.05,0.2) rectangle (17.65,-4.43);
\node[ttl] at (0,0) {(a) Base model stops earlier (positive EOS bias)};
\node[lab] (la1) at (1.55,-0.45) {$b = 0$};
\node[w, right=0.1cm of la1] (a1) {};
\node[ok, right=of a1] (a2) {obj};
\node[w, right=of a2] (a3) {};
\node[ok, right=of a3] (a4) {obj};
\node[w, right=of a4] (a5) {};
\node[bad, right=of a5] (a6) {obj};
\node[w, right=of a6] (a7) {};
\node[ok, right=of a7] (a8) {obj};
\node[w, right=of a8] (a9) {};
\node[bad, right=of a9] (a10) {obj};
\node[w, right=of a10] (a11) {};
\node[eos, right=of a11] (a12) {EOS};
\node[lab] (lb1) at (1.55,-0.9) {$b > 0$};
\node[w, right=0.1cm of lb1] (b1) {};
\node[ok, right=of b1] (b2) {obj};
\node[w, right=of b2] (b3) {};
\node[ok, right=of b3] (b4) {obj};
\node[eos, right=of b4] (b5) {EOS};
\node[wf, right=of b5] (b6) {};
\node[badf, right=of b6] (b7) {obj};
\node[wf, right=of b7] (b8) {};
\node[okf, right=of b8] (b9) {obj};
\node[wf, right=of b9] (b10) {};
\node[badf, right=of b10] (b11) {obj};
\node[lab, align=right] at (1.38,-1.870) {CHAIRi gain\\(pt)};
\draw[black!55] (1.75,-2.220) -- (1.75,-1.476);
\draw[black!55] (1.7,-2.220) -- (1.75,-2.220) node[left=1pt, font=\tiny, text=black!70] {0};
\draw[black!55] (1.7,-1.900) -- (1.75,-1.900) node[left=1pt, font=\tiny, text=black!70] {5};
\draw[black!55] (1.7,-1.580) -- (1.75,-1.580) node[left=1pt, font=\tiny, text=black!70] {10};
\filldraw[fill=blue!35, draw=black!55, line width=0.3pt] (2.220,-2.220) rectangle (2.620,-1.5163);
\filldraw[fill=black!40, draw=black!55, line width=0.3pt] (2.680,-2.220) rectangle (3.080,-1.7886);
\node[font=\scriptsize, anchor=south, inner sep=1pt] at (2.880,-1.7886) {61\%};
\node[font=\scriptsize, anchor=north, inner sep=1pt, text=black!80] at (2.650,-2.240) {RLAIF-V};
\filldraw[fill=blue!35, draw=black!55, line width=0.3pt] (3.870,-2.220) rectangle (4.270,-1.5485);
\filldraw[fill=black!40, draw=black!55, line width=0.3pt] (4.330,-2.220) rectangle (4.730,-1.7014);
\node[font=\scriptsize, anchor=south, inner sep=1pt] at (4.530,-1.7014) {77\%};
\node[font=\scriptsize, anchor=north, inner sep=1pt, text=black!80] at (4.300,-2.240) {CHAIR-DPO};
\filldraw[fill=blue!35, draw=black!55, line width=0.3pt] (5.520,-2.220) rectangle (5.920,-1.6501);
\filldraw[fill=black!40, draw=black!55, line width=0.3pt] (5.980,-2.220) rectangle (6.380,-1.7321);
\node[font=\scriptsize, anchor=south, inner sep=1pt] at (6.180,-1.7321) {86\%};
\node[font=\scriptsize, anchor=north, inner sep=1pt, text=black!80] at (5.950,-2.240) {CSR};
\draw[black!55] (1.75,-2.220) -- (6.45,-2.220);
\filldraw[fill=blue!35, draw=black!55, line width=0.3pt] (6.65,-1.740) rectangle (6.9,-1.580);
\node[anchor=west, font=\scriptsize, inner sep=1pt] at (6.93,-1.660) {PO};
\filldraw[fill=black!40, draw=black!55, line width=0.3pt] (6.65,-2.010) rectangle (6.9,-1.850);
\node[anchor=west, font=\scriptsize, inner sep=1pt] at (6.93,-1.930) {early stop};
\begin{scope}[xshift=8.9cm]
\node[ttl] at (0,0) {(b) PO model writes longer (negative EOS bias)};
\node[lab] (lc1) at (1.55,-0.45) {82 words};
\node[po, right=0.1cm of lc1] (c1) {person};
\node[w, right=of c1] (c2) {};
\node[po, right=of c2] (c3) {bicycle};
\node[w, right=of c3] (c4) {};
\node[eos, right=of c4] (c5) {EOS};
\node[lab] (ld1) at (1.55,-0.9) {132 words};
\node[po, right=0.1cm of ld1] (d1) {person};
\node[w, right=of d1] (d2) {};
\node[po, right=of d2] (d3) {bicycle};
\node[w, right=of d3] (d4) {};
\node[w, right=of d4] (d5) {};
\node[rep, right=of d5] (d6) {people};
\node[w, right=of d6] (d7) {};
\node[rep, right=of d7] (d8) {bicycles};
\node[w, right=of d8] (d9) {};
\node[eos, right=of d9] (d10) {EOS};
\draw[black!45] ([yshift=-0.03cm]d6.south west) -- ([yshift=-0.09cm]d6.south west) -- node[below=-2pt, font=\scriptsize, text=black!65] {repeated categories} ([yshift=-0.09cm]d8.south east) -- ([yshift=-0.03cm]d8.south east);
\node[lab, align=right] at (1.38,-1.870) {Recall change\\(pt)};
\draw[black!55] (1.75,-2.220) -- (1.75,-1.481);
\draw[black!55] (1.7,-2.220) -- (1.75,-2.220) node[left=1pt, font=\tiny, text=black!70] {0};
\draw[black!55] (1.7,-1.990) -- (1.75,-1.990) node[left=1pt, font=\tiny, text=black!70] {5};
\draw[black!55] (1.7,-1.760) -- (1.75,-1.760) node[left=1pt, font=\tiny, text=black!70] {10};
\draw[black!55] (1.7,-1.530) -- (1.75,-1.530) node[left=1pt, font=\tiny, text=black!70] {15};
\filldraw[fill=blue!35, draw=black!55, line width=0.3pt] (2.220,-2.220) rectangle (2.620,-1.6560);
\filldraw[fill=blue!75!black, draw=black!55, line width=0.3pt] (2.680,-2.220) rectangle (3.080,-2.2074);
\node[font=\scriptsize, anchor=south, inner sep=1pt] at (2.880,-2.2074) {2\%};
\node[font=\scriptsize, anchor=north, inner sep=1pt, text=black!80] at (2.650,-2.240) {RLAIF-V};
\filldraw[fill=blue!35, draw=black!55, line width=0.3pt] (3.870,-2.220) rectangle (4.270,-1.5205);
\filldraw[fill=blue!75!black, draw=black!55, line width=0.3pt] (4.330,-2.220) rectangle (4.730,-2.1601);
\node[font=\scriptsize, anchor=south, inner sep=1pt] at (4.530,-2.1601) {9\%};
\node[font=\scriptsize, anchor=north, inner sep=1pt, text=black!80] at (4.300,-2.240) {CHAIR-DPO};
\filldraw[fill=blue!35, draw=black!55, line width=0.3pt] (5.520,-2.220) rectangle (5.920,-1.5426);
\filldraw[fill=blue!75!black, draw=black!55, line width=0.3pt] (5.980,-2.220) rectangle (6.380,-2.1948);
\node[font=\scriptsize, anchor=south, inner sep=1pt] at (6.180,-2.1948) {4\%};
\node[font=\scriptsize, anchor=north, inner sep=1pt, text=black!80] at (5.950,-2.240) {CSR};
\draw[black!55] (1.75,-2.220) -- (6.45,-2.220);
\filldraw[fill=blue!35, draw=black!55, line width=0.3pt] (6.65,-1.740) rectangle (6.9,-1.580);
\node[anchor=west, font=\scriptsize, inner sep=1pt] at (6.93,-1.660) {lost by PO};
\filldraw[fill=blue!75!black, draw=black!55, line width=0.3pt] (6.65,-2.010) rectangle (6.9,-1.850);
\node[anchor=west, font=\scriptsize, inner sep=1pt] at (6.93,-1.930) {regained};
\end{scope}
\draw[black!20] (0,-2.62) -- (17.6,-2.62);
\node[ttl] at (0,-2.87) {(c) Base-guided object-category restoration (no training, no separate visual verifier)};
\node[box, minimum width=6.4cm, minimum height=1.25cm, anchor=north west] (in) at (0.05,-3.13) {};
\node[btl] at (in.north west) {Captions of the same image};
\node[lab] (le1) at (1.6,-3.69) {Base caption};
\node[po, right=0.1cm of le1] (e1) {cake};
\node[bo, right=of e1] (e2) {dining table};
\node[bo, right=of e2] (e3) {person};
\node[bo, right=of e3] (e4) {cup};
\node[bo, right=of e4] (e5) {knife};
\node[lab] (lf1) at (1.6,-4.14) {PO caption};
\node[po, right=0.1cm of lf1] (f1) {cake};
\node[box, minimum width=5.3cm, minimum height=1.25cm, anchor=north west] (sel) at (7.05,-3.13) {};
\node[btl] at (sel.north west) {Keep base-only categories with confidence $> \tau$};
\node[lab] at (8.25,-3.69) {kept};
\node[bo, anchor=west] (g1) at (8.35,-3.69) {dining table};
\node[anchor=west] at (g1.east) {$-0.00$};
\node[bo, anchor=west] (g2) at (10.75,-3.69) {cup};
\node[anchor=west] at (g2.east) {$-0.37$};
\node[lab] at (8.25,-4.14) {dropped};
\node[bof, anchor=west] (h1) at (8.35,-4.14) {person};
\node[anchor=west, text=black!45] at (h1.east) {$-1.40$};
\node[bof, anchor=west] (h2) at (10.75,-4.14) {knife};
\node[anchor=west, text=black!45] at (h2.east) {$-1.55$};
\node[box, minimum width=4.7cm, minimum height=1.25cm, anchor=north west] (out) at (12.9,-3.13) {};
\node[btl] at (out.north west) {PO caption with appended categories};
\node[po, anchor=west] (o1) at (13.1,-3.91) {cake};
\node[bo, right=of o1] (o2) {dining table};
\node[bo, right=of o2] (o3) {cup};
\draw[arr] (in.east) -- (sel.west);
\draw[arr] (sel.east) -- (out.west);
\end{tikzpicture}

%% file: sec/2_method.tex
\needspace{6\baselineskip}
\section{Evaluating PO gains at equal recall}
\label{sec:method}

A hallucination rate can drop because a model mentions fewer objects. We therefore compare each PO model with its base model at matched object recall.
A positive bias on the EOS logit makes the base model stop earlier until it reaches the recall of the PO model (\cref{fig:pipeline}a).
For the diagnosis in \cref{sec:recall}, a negative bias on the EOS logit of the PO model tests whether the PO model returns to the recall of the base model (\cref{fig:pipeline}b).

\subsection{Raw and stopping-controlled gains}
\label{sec:quantities}

Let $\pi_0$ be the base model and $\pi$ a PO model.
For a set of captions, $H$ is the hallucination rate and $R$ is the coverage of ground-truth objects.
On CHAIR, $H$ is CHAIRi and $R$ is object recall.
On AMBER, $H$ is the AMBER CHAIR score and $R$ is Cover.
We compare models at the same $R$ because a comparison of $H$ alone mixes models that mention different numbers of objects.
$R$ measures the share of ground-truth objects that the captions contain, so it is tied to the hallucination metric more directly than the caption length (\cref{tab:share} compares the two).

Let $\pi_0^{b}$ be the base model with a bias $b$ added to the EOS logit at every step.
The stopping curve $S_0(r)$ is the hallucination rate of $\pi_0^{b}$ at the bias $b$ that gives $R(\pi_0^{b}) = r$.
Let $\pi^{\lambda}$ be the PO model with a bias $-\lambda$ on the EOS logit, and let $\lambda^{*}$ be the value that gives $R(\pi^{\lambda^{*}}) = R(\pi_0)$.
The raw gain compares the two models as they are. The stopping-controlled gain compares the PO model with the base model stopped at the recall of the PO model, and the recall-restored gain compares the base model with the PO model extended to the base recall.
We define
\begin{align}
  \Draw &= H(\pi_0) - H(\pi), &
  \DS   &= S_0\big(R(\pi)\big) - H(\pi), \label{eq:gains}\\
  s     &= 1 - \DS / \Draw, &
  \DR   &= H(\pi_0) - H(\pi^{\lambda^{*}}). \nonumber
\end{align}
We call $\Draw$ the raw gain, $\DS$ the stopping-controlled gain, $s$ the stopping share and $\DR$ the recall-restored gain.
The stopping-controlled gain is the part of the gain that the base EOS-bias policy does not reproduce at the recall of the PO model, and the stopping share is the fraction of the raw gain that this policy reproduces.
The stopping share is a ratio of two estimated gains. For a positive raw gain it exceeds 100 percent when $\DS$ is negative and falls below zero when $\DS$ exceeds $\Draw$, and it is unstable when $\Draw$ is small.
The recall-restored gain exists only when the PO model reaches the base recall.

\subsection{Building the EOS stopping curve in one pass}
\label{sec:eoscut}

A simple way to change how long a model writes is to add a constant to the EOS logit at every decoding step.
With greedy decoding the effect of this constant can be computed exactly after generation.
Take a caption generated with bias $b_0$, and let $\ell_t(v)$ be the log-probability of token $v$ at step $t$ of this caption under the unbiased model.
Let $m_t = \max_{v \neq \mathrm{EOS}} \ell_t(v)$.
For any $b \geq b_0$, the greedy output with bias $b$ equals the $b_0$ output cut at the first step $t$ with $\ell_t(\mathrm{EOS}) + b \geq m_t$ and closed with EOS.
Before this step the two outputs have the same prefix and therefore the same logits.
One generation with $b_0 = 0$ thus gives every $b \geq 0$, and one generation with $b_0 = -10$ gives every $b \geq -10$.
A whole stopping curve costs one decoding pass.
This one-pass identity holds for greedy decoding. The equal-recall comparison can be extended to other decoding methods, but their stopping curves need separate decoding runs.

Token-by-token checks against direct generation confirm this identity at positive and negative biases for the base model and every PO model (\appref{B}).
The stopping curve covers only the EOS-bias policy family.
Other policies can lie below it, and a short-answer prompt gives one such point (\appref{C}).

\subsection{Cross-fitted recall matching and bootstrap intervals}
\label{sec:crossfit}

A bias chosen on the evaluation images also fits the noise of those images.
We therefore split the images at random into two halves.
On one half we choose the bias whose recall is closest to the target recall, and on the other half we compare the models with this bias.
We repeat this for five random splits and both directions of each split, and we sum the counts of all ten evaluation halves before computing the rates.
The evaluation (base to PO recall), the diagnosis (PO to base recall) and the restoration threshold on the development set use the same procedure.
On the evaluation halves, the recall gap between the matched base model and the PO model is between $-0.55$ and $+0.18$ points on CHAIR and between $-0.01$ and $+0.04$ points on AMBER.

For the intervals, we resample the images 1,000 times with replacement.
Each draw applies the same image weights to all splits and all models, so the interval of a difference between two models comes directly from the same draws.
We report 95 percent percentile intervals.

%% file: sec/3_setup.tex
\section{Experimental setup}
\label{sec:setup}

\paragraph{Models.}
The base model is LLaVA-1.5-7B~\cite{liu2024llava15}, and we study four public PO checkpoints of this model.
RLAIF-V~\cite{yu2025rlaifv} learns from open-source AI feedback, CHAIR-DPO~\cite{compagnoni2025chairdpo} from a CHAIR-based reward, CSR~\cite{zhou2024csr} from self-rewarding with CLIP-based calibration, and HA-DPO~\cite{zhao2023hadpo} from dispreferred responses with inserted hallucinations.
We leave out three other checkpoints for different reasons.
POVID~\cite{zhou2024povid} passed our validation, but in our protocol it does not lower recall and gains only 0.72 CHAIRi points, so it is outside the condition we study (\appref{D}).
OPA-DPO~\cite{yang2025opadpo} trains LoRA weights in the vision tower, which our validated implementation does not cover.
The TPO checkpoint released with an earlier version of~\cite{he2025tpr} was not validated with our implementation.

\paragraph{Checkpoint validation.}
We ran all models in the HuggingFace (HF) LLaVA implementation and checked the converted weights and the token log-probabilities against the original LLaVA code.
The base model, CHAIR-DPO, CSR and HA-DPO passed these checks.
RLAIF-V had a token log-probability difference above our tolerance, so we also decoded CHAIR-500 for RLAIF-V with the original code.
On 32 images compared token by token, nine captions diverged at near-tie tokens, and four of these divergences fell outside our token-parity criterion.
On CHAIR-500, CHAIRi, recall and the stopping-controlled gain differed by at most 0.48 points between the two implementations, and the interval of the gain was [2.45, 5.76]. This meets our aggregate criterion of at most 0.5 points with an interval above zero.
The rest of the paper uses the HF numbers, and \appref{A} gives all parity tests, checkpoint revisions and library versions.

\paragraph{Benchmarks and scorers.}
CHAIR and AMBER are the main benchmarks, and POPE and MMStar serve as auxiliary checks.
CHAIR-500 uses 500 random images of COCO val2014~\cite{lin2014coco} drawn with seed 0, and AMBER uses its 1,004 generative images.
POPE has 9,000 questions on COCO in the random, popular and adversarial splits, and MMStar~\cite{chen2024mmstar} has 1,500 multiple-choice questions on general ability.
For all models, our CHAIR and AMBER scores equal those of the public chair.py and the official AMBER inference script.
For the base model we obtain an AMBER CHAIR of 7.5 and a Cover of 51.8, close to the 7.6 and 51.7 that CHAIR-DPO reports for LLaVA-1.5-7B~\cite{compagnoni2025chairdpo}.
The held-out set has 500 other COCO val2014 images drawn with seed 1. It comes from the same dataset as CHAIR-500 but shares no image with it.
We use it only to validate restoration with fixed thresholds and to replicate the decomposition.

\begin{table*}[t]
  \centering
  \small
  \setlength{\tabcolsep}{4.6pt}
  \begin{tabular}{l cccccc ccccc}
    \toprule
    & \multicolumn{6}{c}{CHAIR-500} & \multicolumn{5}{c}{AMBER} \\
    \cmidrule(lr){2-7} \cmidrule(lr){8-12}
    Model & CHAIRi$\downarrow$ & $\Hu$$\downarrow$ & Recall$\uparrow$ & F1$\uparrow$ & Words & Obj.
          & CHAIR$\downarrow$ & $\Hu$$\downarrow$ & Cover$\uparrow$ & F1 proxy$\uparrow$ & Obj. \\
    \midrule
    LLaVA-1.5-7B & 14.39 & 26.4 & 80.5 & 76.9 & 89.7 & 3.2 & 7.50 & 12.0 & 51.8 & 65.2 & 4.0 \\
    RLAIF-V      &  3.39 &  7.4 & 68.4 & 78.7 & 93.4 & 2.2 & 3.04 &  5.1 & 50.0 & 65.5 & 3.5 \\
    CHAIR-DPO    &  3.90 &  7.3 & 65.1 & 76.5 & 77.6 & 2.1 & 3.12 &  5.1 & 46.6 & 62.5 & 3.3 \\
    CSR          &  5.49 & 11.4 & 66.0 & 75.7 & 75.1 & 2.2 & 4.41 &  6.7 & 45.1 & 60.8 & 3.3 \\
    HA-DPO       & 12.07 & 21.5 & 76.2 & 77.3 & 83.5 & 2.8 & 6.61 & 10.5 & 49.8 & 64.0 & 3.8 \\
    \bottomrule
  \end{tabular}
  \caption{Standard metrics of the base model and the four PO models. $\Hu$ is the distinct-object hallucination rate. F1 is the set-level object F1 on CHAIR and the distinct-noun F1 proxy on AMBER. Words is the mean number of words per caption. Obj.~is the mean number of distinct object categories per caption on CHAIR and of distinct object nouns on AMBER. All four PO models lower hallucination and coverage together, and they name fewer objects while their captions stay long.}
  \label{tab:standard}
\end{table*}

\begin{table*}[t]
  \centering
  \small
  \setlength{\tabcolsep}{5pt}
  \begin{tabular}{l cccc cccc}
    \toprule
    & \multicolumn{4}{c}{CHAIR-500 (CHAIRi)} & \multicolumn{4}{c}{AMBER (CHAIR)} \\
    \cmidrule(lr){2-5} \cmidrule(lr){6-9}
    & & & \multicolumn{2}{c}{$s$ (\%)} & & & \multicolumn{2}{c}{$s$ (\%)} \\
    \cmidrule(lr){4-5} \cmidrule(lr){8-9}
    Model & $\Draw$ & $\DS$ [95\% CI] & recall & length
          & $\Draw$ & $\DS$ [95\% CI] & recall & length \\
    \midrule
    RLAIF-V   & 10.99 & 4.25 [2.62, 6.07]       & 61 [48, 74]    & 0$^{*}$
              &  4.46 & 3.58 [2.57, 4.57]       & 20 [6, 37]     & 0$^{*}$ \\
    CHAIR-DPO & 10.49 & 2.39 [0.97, 4.13]       & 77 [63, 90]    & 14
              &  4.39 & 1.56 [0.92, 2.30]       & 64 [50, 77]    & 12 \\
    CSR       &  8.90 & 1.28 [$-$0.03, 2.75]    & 86 [72, 100]   & 22
              &  3.09 & $-$0.10 [$-$0.70, 0.55] & 103 [84, 127]  & 19 \\
    HA-DPO$^{\dagger}$ & 2.32 & $-$1.49 [$-$4.40, 1.30] & 164 [$-$432, 897] & 24
              &  0.89 & $-$0.14 [$-$1.04, 0.68] & 115 [45, 469]  & 11 \\
    \bottomrule
  \end{tabular}
  \caption{Part of the PO gain that EOS-only stopping of the base model reproduces. $\Draw$ is the raw gain, $\DS$ the stopping-controlled gain and $s$ the stopping share. Brackets give 95 percent intervals. Recall matching compares the PO model with the base stopping curve at equal recall. Length matching uses the same EOS bias at equal word count. $^{*}$RLAIF-V writes longer captions than the base model, so its length-matched base is the unbiased base model. $^{\dagger}$HA-DPO has a small raw gain, which makes its $s$ unstable, so its $\DS$ is the quantity to interpret.}
  \label{tab:share}
\end{table*}

\paragraph{Metrics.}
We report CHAIRi and CHAIRs. CHAIRi counts every object mention, and many mentions repeat.
On CHAIR 57 to 65 percent of the object mentions repeat a category that the same caption already names, and on AMBER 49 to 54 percent.
We therefore also report the distinct-object hallucination rate $\Hu$.
For each caption we take the set of distinct mentioned objects, and $\Hu$ is the share of these objects that are not in the image, pooled over captions.
$\Hu$ ignores the repeats that CHAIRi counts, so the two rates can rank models differently.
To weigh hallucination against coverage, the set-level object F1 on CHAIR combines the precision $100 - \Hu$ with the recall of distinct ground-truth objects over COCO categories.
On AMBER the same computation uses distinct noun lemmas for the precision and the official Cover for the recall.
The two units differ, so we call the AMBER value the distinct-noun F1 proxy.
The F1 built from $100 - \text{CHAIRi}$ is in \appref{C}.

\paragraph{Decoding and batch size.}
All captions use the prompt ``Please describe this image in detail.'', greedy decoding, at most 512 new tokens, fp16 and the llava\_v1 conversation template.
POPE and MMStar use the short-answer suffixes of the LLaVA evaluation scripts.
Batched fp16 greedy decoding can flip near-tie tokens when the batch size changes, so every run uses one batch size from start to end.
On the development sets, the base model, RLAIF-V, CHAIR-DPO and CSR use batch size 32, and HA-DPO and the negative EOS runs use batch size 16. All held-out runs use batch size 16.
Replacing the base run at batch size 32 with one at batch size 16 changes every reported gain by at most 0.15 points (\appref{J}).

%% file: sec/4_gain.tex
\section{How much of the PO gain can stopping reproduce?}
\label{sec:gain}

\subsection{Lower hallucination, no clear object F1 gain}
\label{sec:f1}
All four PO models lower hallucination, but they also mention fewer object categories (\cref{tab:standard}).
On CHAIR-500 they lower CHAIRi and recall, and on AMBER they lower the AMBER CHAIR and Cover.
They name 2.1 to 2.8 distinct object categories per caption against 3.2 for the base model, while their captions keep 75 to 93 words against 90.
The one-number summaries then disagree.
The distinct-object hallucination rate $\Hu$ is lower for every PO model, by 4.9 to 19.1 points on CHAIR and 1.6 to 6.9 points on AMBER, and all eight intervals exclude zero.
The set-level F1, which uses the same distinct-object unit for precision and recall, changes by $-1.2$ to $+1.8$ points on CHAIR-500 and by $-2.0$ to $+1.2$ points on the held-out set (\appref{C}).
Long \etal~\cite{long2026saying} report a gain for DPO on Qwen2.5-VL in a caption score that also takes the harmonic mean of precision and recall.
For the four public checkpoints of LLaVA-1.5-7B, no PO model gains significantly on either set, and on the held-out set CSR loses 2.0 [$-3.9$, $-0.2$] points.
On AMBER the distinct-noun F1 proxy is unchanged for RLAIF-V and lower for the other three models.
For these models, a lower hallucination rate does not translate into a higher set-level object F1.

\subsection{Base EOS stopping reproduces much of the CHAIRi gain}
Early stopping of the base model reproduces a large share of the CHAIRi gain of RLAIF-V, CHAIR-DPO and CSR (\cref{tab:share}).
On the development sets, the stopping share of these three models is 61, 77 and 86 percent on CHAIR and 20, 64 and 103 percent on AMBER, and all six intervals of $s$ exclude zero.
RLAIF-V has a low share on AMBER because its Cover drops by only 1.8 points.
On the held-out set, with all models decoded at batch size 16, the shares are 58, 61 and 74 percent, and the remaining gains of all three models have intervals above zero (\appref{L}).
At equal recall over the two sets, RLAIF-V keeps 4.3 to 5.0 CHAIRi points, CHAIR-DPO keeps 2.4 to 4.6 points and CSR keeps 1.3 to 2.5 points.
The development interval of CSR, [$-0.03$, 2.75], includes zero and becomes [0.16, 2.82] with a base run at batch size 16 (\appref{J}).
HA-DPO has a small raw gain of 2.32 and 0.89 points, and the intervals of its stopping-controlled gain include zero on the development sets and on the held-out set.

Among the ground-truth categories that the base caption names, those mentioned later are more often missing in the PO captions.
We order all distinct object categories of the base caption by their first mention and split this order into thirds.
RLAIF-V, CHAIR-DPO and CSR then drop 3 to 7 percent of the ground-truth categories in the first third and 36 to 51 percent of those in the last third on both sets.
Hallucinated objects also appear more often late in a caption~\cite{zhou2024lure,han2026egvca}.
The recall-matched base stop also removes some later mentions.
With the absolute token position controlled, we find no significant additional overlap between the categories that PO drops and those that the stop removes (\appref{C}).

We also check whether the remaining gain comes with an image-dependent preference for grounded objects, and whether the POPE changes follow the yes ratio of each model.
In a teacher-forced analysis of object tokens, only RLAIF-V, which keeps the largest gain, shows a large image-dependent preference for grounded objects (\appref{F}).
On POPE, a model that says less answers yes less often, so the analogue of earlier stopping is a shift of the yes ratio.
RLAIF-V is 3.4 to 5.6 accuracy points below the base model over the three POPE splits.
A base model whose yes versus no log-odds are shifted to the yes ratio of each PO model reproduces the accuracy of that PO model within 0.8 points on every split (\appref{G}).
On MMStar we detect no significant accuracy difference for any PO model.

\subsection{Matching length does not match coverage}
LeHaCE~\cite{wei2024lehace} compares models at the same length with a regression over several prompts. We test a single-bias length control, which matches the word count with the same EOS bias.
It reproduces only 0 to 24 percent of the gain (\cref{tab:share}).
The PO models keep long captions and mention fewer objects, so the length-matched base model stays close to the unbiased base model.
RLAIF-V even writes longer captions than the base model, 93.4 against 89.7 words, while it names fewer object categories, 2.2 against 3.2.
At equal recall the PO captions are 1.3 to 1.8 times longer than the captions on the stopping curve (\appref{C}).
CHiP~\cite{fu2025chip} also reports a similar output length with a lower AMBER Cover after PO on LLaVA-1.6.
For these PO models, caption length is a weak proxy for object coverage.

%% file: sec/5_recall.tex
\section{Where the recall goes}
\label{sec:recall}

\begin{table}[t]
  \centering
  \footnotesize
  \setlength{\tabcolsep}{2.6pt}
  \begin{tabular}{l cccc}
    \toprule
    & Recall & Recovered (\%) & Categories & New \\
    Model & drop & CHAIR / AMBER & per caption & (\%) \\
    \midrule
    RLAIF-V   & 12.3 & 2 / 17  & 2.16 $\rightarrow$ 2.17 & 6 [2, 11] \\
    CHAIR-DPO & 15.2 & 9 / 24  & 2.06 $\rightarrow$ 2.13 & 4 [3, 5] \\
    CSR       & 14.7 & 4 / 8   & 2.17 $\rightarrow$ 2.21 & 4 [2, 5] \\
    \midrule
    HA-DPO    & 4.5  & 58 / 76 & 2.83 $\rightarrow$ 3.04 & 8 [6, 11] \\
    \bottomrule
  \end{tabular}
  \caption{Writing longer with a negative EOS bias ($\lambda \leq 10$). Recovered is the largest share of the lost CHAIR recall and AMBER Cover while less than 5 percent of the captions reach the 512-token limit. Categories is the number of distinct object categories per caption on CHAIR from $\lambda = 0$ to that point, against 3.21 for the base model at the same batch size. New is the share of added object mentions that name a new category.}
  \label{tab:debias}
\end{table}

\begin{figure*}[t]
  \centering
  \footnotesize
  \newcommand{\ok}[1]{#1\,\checkmark}
  \newcommand{\no}[1]{#1\,$\times$}
  \newcommand{\qimg}[1]{\IfFileExists{fig/qual_#1.jpg}{\includegraphics[width=\linewidth,height=2.9cm,keepaspectratio]{fig/qual_#1.jpg}}{\fbox{\parbox[c][2.1cm][c]{0.8\linewidth}{\centering\color{red}COCO\\#1}}}}
  \setlength{\tabcolsep}{2pt}
  \renewcommand{\arraystretch}{1.05}
  \begin{minipage}[t]{0.495\linewidth}
    \begin{minipage}[t]{0.29\linewidth}\vspace{0pt}\centering
      \qimg{}\par\vspace{2pt}(a) RLAIF-V
    \end{minipage}\hfill
    \begin{minipage}[t]{0.70\linewidth}\vspace{0pt}
      \begin{tabular}{@{}l >{\raggedright\arraybackslash}p{0.73\linewidth}@{}}
        GT & cake, cup, dining table \\
        Base (88 w.) & \ok{cake}, \ok{dining table}, \no{person}, \ok{cup}, \no{knife} \\
        PO (88 w.) & \ok{cake} \\
        PO, $\lambda{=}10$ & identical to the PO caption \\
        Restored & \ok{cake} $+$ \ok{dining table} ($-0.00$), \ok{cup} ($-0.37$) \\
        Below $\tau$ & \no{person} ($-1.40$), \no{knife} ($-1.55$), $\tau = -1.29$ \\
      \end{tabular}
    \end{minipage}
  \end{minipage}\hfill
  \begin{minipage}[t]{0.495\linewidth}
    \begin{minipage}[t]{0.29\linewidth}\vspace{0pt}\centering
      \qimg{}\par\vspace{2pt}(b) CHAIR-DPO
    \end{minipage}\hfill
    \begin{minipage}[t]{0.70\linewidth}\vspace{0pt}
      \begin{tabular}{@{}l >{\raggedright\arraybackslash}p{0.73\linewidth}@{}}
        GT & bicycle, car, handbag, motorcycle, person, tie \\
        Base (88 w.) & \ok{person}, \ok{handbag}, \ok{bicycle}, \ok{motorcycle} \\
        PO (82 w.) & \ok{person}, \ok{bicycle} \\
        PO, $\lambda{=}10$ & \ok{person}, \ok{bicycle} (132 w.), the added text repeats people, bicycles and signs \\
        Restored & \ok{person}, \ok{bicycle} $+$ \ok{handbag} ($-0.45$), \ok{motorcycle} ($-0.66$) \\
        Below $\tau$ & none, $\tau = -1.71$ \\
      \end{tabular}
    \end{minipage}
  \end{minipage}
  \caption{Two development images of CHAIR-500. Each row lists the object categories that a caption names in order of first mention, with \checkmark\ for a ground-truth category and $\times$ for a hallucinated one, and w.\ is the number of words. PO, $\lambda{=}10$ is the PO model with an EOS bias of $-10$. Restored lists the categories appended to the PO caption with their scores, and $\tau$ is the threshold fixed on the development set. (a) The negative bias leaves the caption unchanged, and restoration adds the two missed categories while the two hallucinated base-only categories stay below $\tau$. (b) The negative bias adds 50 words and no category, and restoration adds the two missed categories that the base caption names. Both examples are successful cases, and \appref{K} shows a failure case.}
  \label{fig:qual}
\end{figure*}

If the recall loss of PO came from the stopping decision, the PO model would recover it by writing longer.
We test this with a negative bias $-\lambda$ on the EOS logit of the PO model.
One generation at $\lambda = 10$ with at most 512 tokens gives every $\lambda \in [0, 10]$ through the cut of \cref{sec:eoscut}, and the cross-fitting of \cref{sec:crossfit} targets the base recall.
The PO runs and the reference base run all use batch size 16.

Extending the PO captions adds little object coverage under these greedy-decoding settings (\cref{tab:debias}).
For RLAIF-V, CHAIR-DPO and CSR, no $\lambda \leq 10$ brings the recall within one point of the base recall in any of the ten calibration folds.
While less than 5 percent of the captions reach the length limit, they recover only 2, 9 and 4 percent of the lost recall on CHAIR and 17, 24 and 8 percent on AMBER.
The captions get longer, but the number of distinct object categories per caption hardly changes.
Only 6, 4 and 4 percent of the added object mentions name a new category, and the other 94 to 96 percent repeat categories that the caption already names.
On captions that reach 100 tokens, the unbiased captions of CHAIR-DPO and CSR keep 85 to 99 percent of the object mentions of the base model between tokens 20 and 100 and name only 39 to 68 percent of its new categories there.
RLAIF-V instead mentions fewer objects in the later window (\appref{I}).
HA-DPO loses only 4.5 recall points and recovers 58 and 76 percent of them, but 60.6 percent of its captions reach the length limit on CHAIR at the matched $\lambda$.
The acceptance rules fixed for this probe and the results of every fold are in \appref{I}.
Within greedy decoding, $\lambda \leq 10$, 512 tokens and our caption prompt, the captions of the three PO models lose object categories that a later stop does not bring back.

The base caption of the same image still names many of these lost categories.
On the CHAIR development set, it names 47 to 48 percent of the ground-truth categories that RLAIF-V, CHAIR-DPO and CSR miss, while the PO captions at $\lambda = 10$ name only 2 to 13 percent.
On AMBER the base caption covers 15 to 18 percent of the missed objects, against 1 to 6 percent at $\lambda = 10$ (\appref{I}).
\Cref{fig:qual} shows two such images, and \cref{sec:restore} brings these categories back from the base caption.

%% file: sec/6_restore.tex
\section{Base-guided object-category restoration}
\label{sec:restore}

\begin{table*}[t]
  \centering
  \footnotesize
  \setlength{\tabcolsep}{1.6pt}
  \begin{tabular}{l ccc ccc cc}
    \toprule
    & \multicolumn{6}{c}{Held-out COCO 500, fixed thresholds} & \multicolumn{2}{c}{Development $\Delta\Hu$} \\
    \cmidrule(lr){2-7} \cmidrule(lr){8-9}
    Model & Recall & $\Hu$ & CHAIRs & $\Delta\Hu$ vs.\ base & Mid vs.\ curve & Vs.\ random & CHAIR & AMBER \\
    \midrule
    LLaVA-1.5-7B & 80.1 & 26.9 & 55.2 & & & & & \\
    \midrule
    RLAIF-V   & 66.4 $\rightarrow$ 79.2 & 8.4 $\rightarrow$ 21.6  & 16.6 $\rightarrow$ 46.4 & 5.24 [3.92, 6.60] & 5.42 [2.74, 8.05]    & 3.77 [2.66, 4.92] & 5.87 [2.93, 8.83]    & 5.66 [4.48, 6.88] \\
    CHAIR-DPO & 64.6 $\rightarrow$ 79.5 & 7.6 $\rightarrow$ 25.0  & 15.2 $\rightarrow$ 52.4 & 1.85 [0.98, 2.77] & 4.36 [2.12, 6.87]    & 1.59 [0.89, 2.37] & 1.97 [0.54, 4.35]    & 2.95 [2.05, 3.88] \\
    CSR       & 64.7 $\rightarrow$ 79.7 & 12.2 $\rightarrow$ 25.8 & 23.0 $\rightarrow$ 54.6 & 1.08 [0.15, 2.05] & 1.91 [$-$0.25, 4.51]    & 1.59 [0.87, 2.31] & 1.42 [$-$0.29, 3.75] & 1.44 [0.60, 2.35] \\
    \midrule
    HA-DPO    & 76.3 $\rightarrow$ 79.5 & 20.9 $\rightarrow$ 23.9 & 40.2 $\rightarrow$ 48.6 & 2.94 [1.46, 4.47] & 1.54 [$-$1.42, 4.25] & 3.09 [2.24, 3.93] & 2.26 [$-$0.53, 3.96] & 0.58 [$-$0.41, 1.80] \\
    \bottomrule
  \end{tabular}
  \caption{Base-guided object-category restoration. Arrows go from the original PO captions to the restored captions, and the first row is the base model. $\Delta\Hu$ is the base value minus the restored value, so positive values favor restoration, and the other differences have the same sign. Mid vs.~curve compares restoration at the midpoint threshold with the base stopping curve refitted on the held-out set. Vs.~random compares with random restoration whose admission probability is fixed on the development set, with the counts averaged over 10 seeds. Its intervals come from the image bootstrap of the seed-averaged counts, and for HA-DPO the random recall is 0.8 points above restoration. The intervals of Mid vs.~curve and of the development set repeat the choice of the bias or threshold in every bootstrap draw. The acceptance criterion covers RLAIF-V, CHAIR-DPO and CSR. Mid vs.~curve and Vs.~random were added after the frozen evaluation.}
  \label{tab:restore}
\end{table*}

\subsection{Selecting base-only categories by confidence}
For each image, the candidates are the object categories that the base caption names and the PO caption does not.
The score of a candidate is the log-probability of the first token of its first mention in the base caption, taken from the greedy pass of the base model.
For each candidate with a score above $\tau$, we append the sentence ``There is also a \{object\} in the image.'' to the PO caption (\cref{fig:pipeline}c).
The selection uses no ground truth and no separate visual verifier, and the score ranks the base-only categories by the confidence of the base model.
The appended sentence is a template, so we evaluate the restoration of object categories and make no claim on caption fluency.

We select $\tau$ for each PO model and each benchmark so that the restored captions reach the base recall, since CHAIR and AMBER extract objects differently.
For the development results we choose it with the cross-fitting of \cref{sec:crossfit}. For the held-out set we choose one $\tau$ per model, $\tau_0$, on the whole CHAIR development set and apply it unchanged.
Both choices use the ground-truth objects of the development images only.
A second threshold aims at the midpoint between the base and PO recalls.

Four controls test the two parts of the method, the confidence score and the PO caption as the starting point.
Random restoration tests the confidence score. It appends each base-only candidate with one probability, fixed on the development set in the same way as $\tau$ and applied to the held-out set with 10 seeds. Union restoration appends all base-only candidates. Base-only filtering keeps the base caption and removes its low-confidence objects, and the base stopping curve lets the base model stop earlier. These two test whether the base model alone reaches the same trade-off.
We designed restoration after seeing the development results, and we then fixed the thresholds and one acceptance criterion before generating the held-out set.
The criterion asks that RLAIF-V, CHAIR-DPO and CSR all reach a recall within one point of the base recall with an $\Hu$ reduction whose interval lies above zero.
The random and union controls, the comparison with the stopping curve at $\tau_0$, the bootstrap that repeats the calibration of the stopping bias, the sweep of $\tau$ on the held-out set and the candidate-source controls were added after this evaluation (\appref{K}).

\subsection{Recall recovery and matched-recall comparison}
With the thresholds fixed on the development set, restoration brings RLAIF-V, CHAIR-DPO and CSR close to the base recall on the held-out set (\cref{tab:restore}).
Their restored recall is 79.2, 79.5 and 79.7 against 80.1 for the base model, and their $\Hu$ is 21.6, 25.0 and 25.8 against 26.9.
Restoration thus lowers $\Hu$ below the base model by 5.24 [3.92, 6.60], 1.85 [0.98, 2.77] and 1.08 [0.15, 2.05] points, and all three models meet the acceptance criterion.
HA-DPO is not covered by the criterion, which we fixed for the three models with a large recall drop. It also comes within one point of the base recall, and its $\Hu$ is 2.94 [1.46, 4.47] points below the base model.
The development set gives the same direction on CHAIR and AMBER.
With the threshold choice repeated inside every bootstrap draw, only CSR on CHAIR has an interval that includes zero.

Restoration keeps only part of the PO gain, and $\Hu$ and CHAIRs move most of the way back to the base level.
In return, the set-level F1 rises over the PO captions by 1.80 [0.45, 3.31], 1.19 [$-0.38$, 2.96] and 2.38 [0.80, 4.02] points, and only the interval of CHAIR-DPO includes zero.

The appended categories are often wrong.
On the held-out set, the base caption names 46 to 48 percent of the ground-truth categories that the three PO captions miss, but only 38.5 to 41.8 percent of all base-only candidates are such recoverable categories.
The appended categories have a higher precision of 41.4 to 44.8 percent, and the correct ones cover 79 to 91 percent of the recoverable categories (\appref{K}).
More than half of the appended categories are thus hallucinated. The restored captions still stay below the base model in $\Hu$ because the PO captions they start from are precise.

The recovered categories are mostly those that writing longer does not bring back.
In a post-hoc analysis of the development runs, 92 to 98 percent of the base-only ground-truth categories stay missing at the negative EOS bias of \cref{sec:recall} with the largest recall under the length limit, and restoration with $\tau_0$ fills 81 to 92 percent of them (\appref{K}).

The acceptance criterion compares restoration with the unbiased base model. A stricter post-hoc comparison uses the base EOS stopping curve at matched recall, and it gives a more limited result.
We refit this curve on the held-out set, and every bootstrap draw repeats the calibration of the stopping bias.
At the midpoint threshold fixed on the development set, restoration has an $\Hu$ lower than the curve by 5.42, 4.36 and 1.91 points, and the intervals of RLAIF-V and CHAIR-DPO exclude zero.
At $\tau_0$, which aims at the base recall, the restored recall is 0.4 to 1.0 points below the base recall, where the curve drops fast. The gap to the curve is 3.10 [0.31, 5.21] for RLAIF-V, 0.18 [$-1.68$, 1.58] for CHAIR-DPO and $-0.28$ [$-2.38$, 1.10] for CSR.
Part of the gap to the base model thus comes from the lower recall.
In a sweep of $\tau_0 \pm 0.5$, wherever the restored recall is below the base recall, RLAIF-V stays below the curve by 2.5 to 3.4 points with intervals above zero, and CHAIR-DPO and CSR stay within 1.1 points of it with intervals that include zero. Thresholds that push the recall above the base recall bring $\Hu$ to the base level or above for these two models.
Base-only filtering never exceeds the base recall and is worse than the base stopping curve at the midpoint recall on both development sets (\appref{K}).

Random and union controls test whether selecting candidates by base confidence is useful.
On the held-out set, random restoration reaches a recall 0.26 to 0.35 points above restoration. Its $\Hu$ is higher by 1.59 to 3.77 points, with intervals above zero (\cref{tab:restore}).
Union restoration reaches a recall of 81.0 to 82.6, but its $\Hu$ of 27.5 to 28.4 is above the base model, and restoration has a set-level F1 0.68 to 1.77 points higher than the union with intervals above zero (\appref{K}).
Among the base-only candidates, the base confidence therefore selects correct categories better than chance. Its AUC of 0.68 to 0.69 drops to 0.52 to 0.54 when the image is swapped, so the selection uses the image (\appref{K}).
With image-wise count targets taken from the frozen restoration, candidates from the base captions of other images, a permutation of the appended categories and priors estimated on the development set reach a precision of only 5 to 20 percent, against 41 to 45 percent for the base-only candidates of the same image (\appref{K}).

%% file: sec/7_discussion.tex
\section{Discussion}
\label{sec:discussion}

\paragraph{Reporting hallucination with coverage.}
Hallucination gains of PO models should be reported together with recall or Cover and with the stopping-controlled gain.
When a model lowers hallucination and recall together, CHAIRi and the set-level F1 can point in different directions, as for CSR on the held-out set.
At equal recall, the hallucination rate of the PO model is compared with that of the base EOS-bias policy, without an exchange rate between the two metrics.
POPE accuracy should be reported with the yes ratio.

\paragraph{Preserving object coverage during PO.}
For the three PO models with a large recall drop, extending the captions adds few new object categories. Future PO training should therefore target grounded category coverage and keep the hallucination reduction that remains at matched recall. Preference pairs at similar recall and coverage-aware rewards that favor verified objects are two candidates to test by set-level F1 and hallucination at matched recall. Base-only candidates need verification, since only 38.5 to 41.8 percent of them are correct on the held-out set.

\paragraph{Relation to prior mitigation methods.}
Work on stopping and length shows that shorter outputs can lower hallucination scores.
Yue \etal~\cite{yue2024less} reduce hallucination by training the EOS decision, Fazli \etal~\cite{fazli2026safe} show that inference-time methods trade informativeness for lower hallucination scores, and GACD~\cite{wang2026gacd} reports a small recall loss as an advantage.
LeHaCE~\cite{wei2024lehace} controls length, and for the PO checkpoints that we study, recall is a more direct control for object hallucination than caption length.
We measure this trade-off for trained PO checkpoints with a counterfactual of the base model, where early stopping reproduces most of the CHAIRi gain of three PO models.
Bu and Gao~\cite{bu2026buy} also use a recall-matched control with truncated base captions.
They measure how many hallucinated-object removals a frozen probe reproduces beyond position-matched random deletion, and RLAIF-V has the lowest accounted share on COCO.
CHiP~\cite{fu2025chip}, Long \etal~\cite{long2026saying} and Alqahtani \etal~\cite{alqahtani2026beyond} report the coverage loss of PO on other backbones or training setups.
In addition to the matched-recall comparison, we test the reverse direction with the same EOS bias, where a later stop of the PO models recovers little of the lost recall, and we measure how many of the missed categories the base caption of the same image recovers and at what hallucination cost.
Other inference-time methods use reward-guided decoding (MRGD~\cite{manas2025mrgd}), image-dependent claim admission (EGVCA~\cite{han2026egvca}), self-validation of object existence in sampled captions~\cite{liu2026selfval} or local image crops (VIGOR~\cite{li2026vigor}).
PatchGate~\cite{ko2026patchgate} and ReflectCAP~\cite{min2026reflectcap} recover omitted objects without training, using internal patch evidence or reflection notes.
Our restoration uses the base caption of the same image to recover categories that PO misses, without additional training or a separate visual verifier.
It needs one more greedy pass of the base model, and only 41 to 45 percent of the categories it appends on the held-out set are correct.
Si \etal~\cite{si2025omission} report that reducing omission tends to raise fabrication, and restoration shows this cost in $\Hu$.
Training-time grounding methods such as V-DPO~\cite{xie2024vdpo} and C$^2$-DPO~\cite{ko2026c2dpo} train PO models for visual or context grounding, and the recall-controlled evaluation applies to them as well.

\paragraph{Scope and limitations.}
We study one backbone, LLaVA-1.5-7B, with greedy decoding and one detailed-caption prompt, and our two COCO sets are disjoint but come from the same dataset.
Attribute and relation hallucination and the checkpoints we could not validate are outside this study.
The stopping curve belongs to the EOS-bias policy family, so recall matching controls only the part that EOS-only stopping reproduces, and caption length and object density still differ at equal recall (\appref{C}).
The stopping share is a reproduction ratio, and we make no causal claim on the training effect.
All intervals are nominal 95 percent intervals without a correction for multiple comparisons.
The position, discovery and candidate-source analyses of \appref{C}, \appref{I} and \appref{K} reuse the existing runs and are post-hoc analyses.
The hallucinated categories that restoration appends come mostly from the late part of the base caption.
On the held-out set, 63 to 78 percent of them are categories that the recall-matched stop also removes, and such candidates have a precision of 33 to 39 percent against 54 to 59 percent for the other candidates (\appref{K}).
We do not compare restoration with other training-free restoration methods such as PatchGate under a shared protocol.
Restoration covers only the object categories or noun lemmas that each benchmark recognizes.

\paragraph{Conclusion.}
For RLAIF-V, CHAIR-DPO and CSR under greedy decoding, early stopping of the base model reproduces 58 to 86 percent of the CHAIRi gain at equal recall, while a later stop of the PO models recovers only 2 to 9 percent of their lost CHAIR recall.
Together, recall-matched evaluation and the negative EOS probe distinguish these two effects, and base-guided restoration recovers most of the lost recall from the base caption without training at the cost of part of the PO gain.

%% file: sec/8_llm.tex
\section*{LLM Usage Statement}
We used Claude (Anthropic) and ChatGPT (OpenAI) to assist with research planning, implementation and debugging of analysis code, inspection of intermediate results, and drafting and editing the manuscript and LaTeX source.
The reported benchmark scores were checked against official evaluation implementations where available, and the derived statistics were checked against saved outputs.
The author takes full responsibility for the methodology, results and content of the paper.

%% file: sec/S1_appendix_A_F.tex
\section{Checkpoint validation}
\label{app:A}

\paragraph{Implementation.}
All models run in the HuggingFace (HF) \texttt{LlavaForConditionalGeneration} built from \texttt{llava-hf/llava-1.5-7b-hf}, the official conversion of \texttt{liuhaotian/llava-v1.5-7b}.
The released PO weights use the original LLaVA format.
We convert full checkpoints by renaming keys, and we merge LoRA adapters into the weights by hand.
As in the original LLaVA builder, the vision-tower tensors of a checkpoint are used only when its config sets \texttt{unfreeze\_mm\_vision\_tower}.
Otherwise the model uses the openai CLIP ViT-L/14-336 tower, which is bit-identical to the llava-hf tower in our mapping self-test.
OPA-DPO trains LoRA weights in the vision tower, which this check does not cover, so we leave it out.
\Cref{tab:revisions} lists the repositories and revisions of the four PO models in the main study.
All runs use transformers 4.51.3, fp16 and the \texttt{llava\_v1} conversation template.

\begin{table}[h]
  \centering
  \scriptsize
  \setlength{\tabcolsep}{2pt}
  \begin{tabular}{lll}
    \toprule
    Model & Repository & Revision \\
    \midrule
    RLAIF-V   & \texttt{openbmb/RLAIF-V-7B} & \texttt{90e8e73} \\
    CHAIR-DPO & \texttt{aimagelab/CHAIR-DPO\_LLaVA-1.5-7B} & \texttt{bd9ab77} \\
    CSR       & \texttt{charlesdj/CSR\_LLaVA\_1.5\_7b\_2Iteration} & \texttt{7ebdd0b} \\
    HA-DPO    & \texttt{juliozhao/hadpo-llava-1.5} & \texttt{72b1bcf} \\
    \bottomrule
  \end{tabular}
  \caption{PO checkpoint repositories and revisions used in this study. HA-DPO is a LoRA adapter.}
  \label{tab:revisions}
\end{table}

\paragraph{Weight check.}
For each checkpoint we rebuild the language model with the HF loaders, using \texttt{LlamaForCausalLM.from\_pretrained} for full checkpoints and a PEFT merge for LoRA adapters.
The rebuilt language model and the projector match our converted model bit for bit at fp16.

\paragraph{Parity with the original code.}
On 4 images with 32 tokens, the prompt token ids and pixel values of our pipeline and of the original LLaVA code match exactly.
The token log-probability difference is at most 0.064, and the argmax tokens agree at 127 or 128 of 128 positions.
On 16 images with up to 256 tokens, the base model, CHAIR-DPO, CSR and HA-DPO have a log-probability difference of at most 0.066, and their greedy captions are identical for 14 to 16 of 16 images.
RLAIF-V has a difference of 0.25, above our threshold of 0.1, an argmax agreement of 99.7 percent and identical captions for 10 of 16 images.

\paragraph{RLAIF-V with the original code.}
We therefore decoded RLAIF-V with the original LLaVA code, installed from the main branch on the day of the run (head \texttt{c121f04}).
The \texttt{generate()} call of the original code fails with transformers 4.51 because of the \texttt{cache\_position} argument, so we wrap the original forward pass in a greedy loop with a KV cache.
On the first image this loop agrees with a forward pass without cache at all 126 positions.
We fixed two acceptance criteria before the run.
Token parity requires that every divergence has a margin between 0 and 0.1 nat for both implementations.
Aggregate robustness requires that CHAIRi, recall and the stopping-controlled gain on CHAIR-500 move by at most 0.5 points and that the interval of the gain excludes zero.

In step 1, the HF model and the original code both decode the same 32 images at batch size 1.
They give identical captions for 23 images.
\Cref{tab:diverge} lists the nine divergences. All margins are below 0.1 nat in absolute value, and four are negative, so token parity fails.
In step 2, the original code decodes all of CHAIR-500 at batch size 1, and 377 of 500 captions equal our HF captions at batch size 32.
\Cref{tab:origagg} compares the aggregate results, which meet the aggregate robustness criterion.
The main text uses the HF numbers.

\begin{table}[h]
  \centering
  \footnotesize
  \setlength{\tabcolsep}{3.5pt}
  \begin{tabular}{rrlrr}
    \toprule
    COCO id & Position & Diverging tokens & Original & HF \\
    \midrule
    7276  & 105 & en / giving & $-$0.016 & 0.016 \\
    8204  & 14  & building / tall & 0.031 & 0.024 \\
    8953  & 84  & interaction / view & 0.055 & 0.001 \\
    11374 & 32  & with / appearing & 0.031 & 0.006 \\
    13004 & 64  & presentation / plate & 0.016 & $-$0.014 \\
    13783 & 22  & seems / appears & 0.094 & $-$0.014 \\
    21088 & 46  & trees / hills & 0.031 & 0.054 \\
    31050 & 98  & and / or & 0.000 & 0.046 \\
    33499 & 9   & gold / intr & $-$0.031 & 0.046 \\
    \bottomrule
  \end{tabular}
  \caption{The nine divergences of RLAIF-V between the original code and the HF model at batch size 1. Original and HF give the log-probability margin (nat) by which each implementation prefers its own token at the first divergent position.}
  \label{tab:diverge}
\end{table}

\begin{table}[h]
  \centering
  \footnotesize
  \setlength{\tabcolsep}{4pt}
  \begin{tabular}{lcc}
    \toprule
    RLAIF-V on CHAIR-500 & HF, batch 32 & Original, batch 1 \\
    \midrule
    CHAIRi & 3.39 & 3.42 \\
    Recall & 68.42 & 67.95 \\
    $\Hu$ & 7.41 & 7.55 \\
    Set-level F1 & 78.69 & 78.33 \\
    Words & 93.37 & 93.49 \\
    $\DS$ [95\% CI] & 4.25 [2.62, 6.07] & 4.15 [2.45, 5.76] \\
    Stopping share & 61\% & 62\% \\
    \bottomrule
  \end{tabular}
  \caption{RLAIF-V on CHAIR-500 with our HF pipeline and with the original LLaVA code.}
  \label{tab:origagg}
\end{table}

\paragraph{Scorers.}
Our CHAIR scorer follows the public \texttt{chair.py} with nltk tokenization, the Pattern singularizer, the double-word list and the synonym table.
Our AMBER scorer re-implements the generative branch of the official \texttt{inference.py} per caption.
For all models, our aggregate CHAIR and AMBER scores equal those of the public scripts. On the restored captions of \cref{app:K}, the official \texttt{chair.py} and our scorer agree within 0.08 points on every aggregate metric, and the AMBER CHAIR, Cover, Hal and Cog of both AMBER scorers are identical.
Exact CHAIR scores need nltk 3.9.1. With nltk 3.10.4, one RLAIF-V caption of the held-out set is tokenized differently, and the held-out $\Hu$ of RLAIF-V restoration becomes 21.69 instead of 21.64.

\section{Exactness of the post-hoc EOS cut}
\label{app:B}

\paragraph{Claim.}
Let a greedy caption be decoded with an EOS bias $b_0$.
Let $\ell_t(v)$ be the log-probability of token $v$ at step $t$ of this caption under the unbiased model, and let $m_t = \max_{v \neq \mathrm{EOS}} \ell_t(v)$.
For any $b \geq b_0$, the greedy caption with bias $b$ equals the $b_0$ caption cut at the first step $t$ with $\ell_t(\mathrm{EOS}) + b \geq m_t$ and closed with EOS.

\paragraph{Argument.}
Before this step, the $b_0$ caption picks a non-EOS token $v_t$ with $\ell_t(v_t) = m_t > \ell_t(\mathrm{EOS}) + b \geq \ell_t(\mathrm{EOS}) + b_0$.
The bias $b$ changes only the EOS logit, so the greedy choice with bias $b$ is the same $v_t$, and the two prefixes stay equal.
At the cut step, EOS wins under bias $b$.
In the code we compare $\ell_t(\mathrm{EOS}) + b$ and $m_t$ in float32, as the decoder does, so that ties resolve in the same way.

\paragraph{Checks.}
For positive biases, we decoded 16 CHAIR images directly with $b \in \{1, 2, 3, 4, 6\}$, and the post-hoc cut of the unbiased run equals the direct decoding for 16 of 16 captions at every bias.
As an additional check over a wider range of negative biases, we used a small randomly initialized LLaVA model. With $b_0 = -6$ and $b_0 = -10$, the post-hoc cut matches direct decoding at 61 and 77 biases.
On our GPU runs with $b_0 = -10$, the cut matches direct decoding for 16 of 16 images at five biases $\{-8, -4, -2, -1, 0\}$ of the base model and at three biases $\{-8, -4, 0\}$ of the first batch of every PO run, each at the batch size of its run.
The stored greedy captions at batch size 32 and the $\lambda = 0$ cut of the de-biasing runs at batch size 16 agree for 397 to 500 of 500 captions, and the difference comes from the batch layout (\cref{app:J}).
The identity holds for greedy decoding only.

\section{Stopping curve family and auxiliary matching}
\label{app:C}

\paragraph{Policy family.}
The stopping curve $S_0(r)$ covers one policy family, the base model with a single global EOS bias, and other policies can reach a lower CHAIRi at the same recall.
\Cref{tab:aux} compares it with other ways to move the base model to the recall of a PO model.
The randomized mixture of two adjacent EOS biases hits the calibration recall exactly and gives the same share as the single bias.
The short-answer policy mixes the base captions with base captions for ``Please describe this image briefly.'' (``Describe this image briefly.'' on AMBER).
The short-answer captions have 74 and 64 words on average, and their recall of 79.04 on CHAIR-500 and Cover of 50.56 on AMBER are above those of every PO model.
This policy thus stays at the short-answer point, and \cref{tab:aux} reports the share of that point.
At the same recall, the stopping curve on all images has a CHAIRi of 13.15 against 11.46 for the short-answer point on CHAIR-500, a difference of 1.69 [$-0.63$, 3.48], and an AMBER CHAIR of 6.95 against 6.13, a difference of 0.83 [0.09, 1.72].
The short-answer point thus lies below the stopping curve, and the interval excludes zero on AMBER.
The two oracle rows truncate each base caption to the word count or to the object-mention count of the PO caption of the same image. They look at the PO output, so they are diagnostics and do not serve as a matching axis.
The exploratory frontier takes the best point over EOS biases, first-$k$-sentence truncations and the short-answer prompt on the evaluation images.

\begin{table}[h]
  \centering
  \footnotesize
  \setlength{\tabcolsep}{1.8pt}
  \begin{tabular}{l ccc ccc}
    \toprule
    & \multicolumn{3}{c}{CHAIR-500} & \multicolumn{3}{c}{AMBER} \\
    \cmidrule(lr){2-4} \cmidrule(lr){5-7}
    Policy & RLAIF-V & C-DPO & CSR & RLAIF-V & C-DPO & CSR \\
    \midrule
    EOS bias (main) & 61 & 77 & 86 & 20 & 64 & 103 \\
    EOS mixture & 61 & 77 & 86 & 20 & 65 & 103 \\
    Short-answer prompt & 27 & 28 & 33 & 31 & 31 & 44 \\
    Oracle words & 6 & 19 & 28 & 7 & 20 & 31 \\
    Oracle objects & 52 & 52 & 57 & 28 & 37 & 52 \\
    Exploratory frontier & 59 & 77 & 90 & 31 & 65 & 104 \\
    \bottomrule
  \end{tabular}
  \caption{Share (\%) of the raw gain that each base-model policy reproduces on the development sets. C-DPO is CHAIR-DPO. Only the first two rows are cross-fitted.}
  \label{tab:aux}
\end{table}

\paragraph{Length and density at equal recall.}
\Cref{tab:equalrecall} compares each PO model with the stopping-curve captions at the same recall.
The PO captions are 1.3 to 1.8 times longer and mention more objects per caption, so recall matching leaves caption length and object density unmatched.

\begin{table}[h]
  \centering
  \footnotesize
  \setlength{\tabcolsep}{1.9pt}
  \begin{tabular}{l ccccc}
    \toprule
    Model & Recall & CHAIRi & CHAIRs & Objects & Words \\
    \midrule
    RLAIF-V   & 68.4 / 67.9 &  3.4 / 7.7  & 14.8 / 22.4 & 5.5 / 4.6 & 93.4 / 51.8 \\
    CHAIR-DPO & 65.1 / 65.0 &  3.9 / 6.3  & 14.0 / 18.6 & 5.9 / 4.2 & 77.6 / 47.6 \\
    CSR       & 66.0 / 66.2 &  5.5 / 6.8  & 21.0 / 20.0 & 6.2 / 4.3 & 75.1 / 49.2 \\
    HA-DPO    & 76.2 / 76.4 & 12.1 / 10.6 & 38.6 / 34.6 & 7.2 / 5.8 & 83.5 / 66.5 \\
    \bottomrule
  \end{tabular}
  \caption{PO captions and base stopping-curve captions at the same recall on CHAIR-500. Each cell gives the PO value and the curve value. Objects is the number of object mentions per caption. Curve values are pooled over the ten evaluation halves.}
  \label{tab:equalrecall}
\end{table}

\paragraph{F1 from CHAIRi.}
The F1 that combines $100 - \text{CHAIRi}$ with recall counts repeated mentions in the precision.
Against the base model it changes by $-2.86$ [$-4.30$, $-1.29$] for RLAIF-V, $-5.32$ [$-6.95$, $-3.55$] for CHAIR-DPO and $-5.22$ [$-6.74$, $-3.59$] for CSR on CHAIR-500, and by $-0.42$ [$-1.31$, 0.41], $-3.47$ [$-4.24$, $-2.73$] and $-5.14$ [$-5.89$, $-4.38$] on AMBER.
This F1 thus also shows no gain for these PO models.

\paragraph{Set-level F1 by model.}
The ranges in Sec.~\ref{sec:f1} come from the following changes against the base model. The set-level F1 changes by $+1.8$ [$-0.01$, 4.0] for RLAIF-V, $-0.4$ [$-2.4$, 1.8] for CHAIR-DPO, $-1.2$ [$-3.1$, 0.9] for CSR and $+0.4$ [$-1.0$, 1.9] for HA-DPO on CHAIR-500, and by $+0.5$ [$-1.3$, 2.2], $-0.5$ [$-2.4$, 1.3], $-2.0$ [$-3.9$, $-0.2$] and $+1.2$ [$-0.2$, 2.4] on the held-out set.

\begin{table*}[t]
  \centering
  \footnotesize
  \setlength{\tabcolsep}{4pt}
  \begin{tabular}{ll ccc ccc}
    \toprule
    & & \multicolumn{3}{c}{Dropped by PO (\%)} & \multicolumn{3}{c}{Overlap with the matched stop over a null} \\
    \cmidrule(lr){3-5} \cmidrule(lr){6-8}
    Set & Model & First third & Middle third & Last third & Null A & Null C, rank bins & Null C, token bins \\
    \midrule
    Dev. & RLAIF-V & 6 & 23 & 36 & 1.69 [1.56, 1.90] & 1.14 [1.06, 1.25] & 1.06 [0.99, 1.14] \\
     & CHAIR-DPO & 5 & 23 & 46 & 1.69 [1.56, 1.85] & 1.11 [1.05, 1.17] & 1.01 [0.96, 1.05] \\
     & CSR & 3 & 20 & 48 & 1.79 [1.65, 1.95] & 1.11 [1.05, 1.15] & 1.00 [0.96, 1.04] \\
    \midrule
    Held-out & RLAIF-V & 7 & 27 & 38 & 1.77 [1.63, 1.94] & 1.15 [1.07, 1.24] & 1.05 [0.97, 1.12] \\
     & CHAIR-DPO & 5 & 22 & 48 & 1.82 [1.68, 1.96] & 1.10 [1.04, 1.17] & 0.98 [0.94, 1.03] \\
     & CSR & 4 & 20 & 51 & 1.88 [1.73, 2.04] & 1.13 [1.07, 1.19] & 1.00 [0.96, 1.05] \\
    \bottomrule
  \end{tabular}
  \caption{Position of the ground-truth categories that the PO caption drops. The thirds split the relative first-mention position of all distinct object categories in the base caption, and the drop rates count only the ground-truth categories among them. The overlap counts the dropped categories that the recall-matched base stop also removes, averaged over the five cross-fitting splits, and each column gives its ratio to a null. Null A draws the dropped categories at random within each image. Null C keeps the number of dropped categories of each image and weights each category by the drop probability of its position bin, estimated on the development set. Rank bins split this relative position into five equal parts and token bins are 20 tokens wide. Intervals come from 1,000 image bootstrap draws that repeat the calibration of the stop, and the Null C intervals are conditional on the estimated probabilities.}
  \label{tab:position}
\end{table*}

\paragraph{Position of the dropped categories.}
We test where the ground-truth categories that PO drops sit in the base caption.
For each image we order the $n$ distinct object categories of the base caption, ground-truth and hallucinated, by their first mention, and the category at rank $r = 0, \dots, n-1$ gets the relative position $r / \max(n-1, 1)$.
We split this relative position into thirds and count, for the ground-truth categories in each third, the share that the PO caption drops.
\Cref{tab:position} gives these shares and the overlap of the drops with the categories that the recall-matched stop of Sec.~\ref{sec:crossfit} removes, and every bootstrap draw repeats the calibration of the stop.
Null A draws the dropped categories at random within each image.
Null B keeps the drop rate of each stratum of position bin and number of base ground-truth categories.
Null C keeps the number of dropped categories of each image and weights each category by the drop probability of its position bin, which we estimate on the development set by two-fold cross-fitting and fix for the held-out set.
On both sets the drop rate rises from 3 to 7 percent in the first third to 36 to 51 percent in the last third.
The overlap is 1.69 to 1.88 times Null A and 1.10 to 1.15 times Null C with rank bins, and all these intervals lie above one.
With token bins the ratio to Null C is 0.98 to 1.06 and all six intervals include one.
Null B gives 1.17 to 1.37 with rank bins and 0.98 to 1.06 with token bins.
The ratio to Null C stays at 1.10 to 1.15 for 3 and 10 rank bins and at 0.98 to 1.06 for 10 and 40 token bins, and a linear probability model with position dummies gives the same pattern.
With the absolute token position controlled we thus find no significant additional overlap between the drops of PO and the matched stop.

\section{POVID}
\label{app:D}

POVID passed the same validation as the other checkpoints.
In our protocol it does not lower recall, so it is outside the condition we study.
On CHAIR-500 it has a CHAIRi of 13.67 against 14.39 for the base model, a recall of 81.4 against 80.5, a CHAIRs of 58.0 against 47.2 and 96.9 words against 89.7.
The raw gain of 0.72 [$-0.70$, 2.04] needs no earlier stop, so the matched EOS bias is 0.
On AMBER, its CHAIR of 9.80 is above the 7.50 of the base model, with a Cover of 52.4.
On POPE it raises the yes ratio from 40.6, 41.8 and 43.9 to 43.2, 45.6 and 48.3 percent. Against the yes-matched base model its accuracy differs by 0.16 [$-0.19$, 0.48], $-0.85$ [$-1.31$, $-0.45$] and $-0.37$ [$-0.87$, 0.12] points.
On MMStar it is 1.33 [0.07, 2.73] points above the base model, and its teacher-forced image dependence is $-0.62$ [$-1.13$, $-0.14$] (\cref{app:F}).

\section{PO combined with VCD}
\label{app:E}

As an exploratory check, we combined greedy decoding with visual contrastive decoding (VCD) with $\alpha = 1$, $\beta = 0.1$, 500 noise steps and a fixed noise seed per image.
On CHAIR-500, VCD raises the CHAIRi of the base model from 14.39 to 14.98 at a recall of 80.5 and that of RLAIF-V from 3.39 to 5.06 at a recall of 68.1.
With RLAIF-V as the reference model, RLAIF-V with VCD lies 1.75 [0.93, 2.52] CHAIRi points above the stopping curve of RLAIF-V at the same recall. For the base model, the same comparison gives 0.76 [$-0.73$, 2.60] points and its interval includes zero.

\section{Teacher-forced object token analysis}
\label{app:F}

\begin{table*}[t]
  \centering
  \footnotesize
  \setlength{\tabcolsep}{6pt}
  \begin{tabular}{l cc c cc}
    \toprule
    Model & Sel.\ true [95\% CI] & Sel.\ swap & Dependence [95\% CI] & $d_{\mathrm{g}}$ & $d_{\mathrm{h}}$ \\
    \midrule
    RLAIF-V   & 4.21 [3.64, 4.89] & 0.85 & 3.36 [2.71, 3.98] & $-$2.70 & $-$6.92 \\
    CHAIR-DPO & 0.52 [0.46, 0.58] & 0.52 & $-$0.00 [$-$0.08, 0.07] & $-$0.03 & $-$0.55 \\
    CSR       & 0.32 [0.27, 0.37] & 0.51 & $-$0.20 [$-$0.28, $-$0.12] & 0.05 & $-$0.27 \\
    HA-DPO    & 0.08 [0.06, 0.11] & 0.15 & $-$0.06 [$-$0.10, $-$0.03] & $-$0.03 & $-$0.11 \\
    POVID     & 0.42 [0.33, 0.53] & 1.04 & $-$0.62 [$-$1.13, $-$0.14] & 0.12 & $-$0.30 \\
    \bottomrule
  \end{tabular}
  \caption{Teacher-forced selectivity on the base captions of CHAIR-500, model minus base, in nat. Dependence is the selectivity with the true image minus that with a swapped image.}
  \label{tab:tf}
\end{table*}

We fix the greedy base caption of each CHAIR-500 image and score it with each model.
At the first token of each object word, we take the log-probability difference between the model and the base model, $d$.
We average $d$ over grounded mentions ($d_{\mathrm{g}}$) and over hallucinated mentions ($d_{\mathrm{h}}$), and the selectivity is $d_{\mathrm{g}} - d_{\mathrm{h}}$.
We repeat the scoring with the image of another CHAIR-500 image in the prompt.
The image dependence is the selectivity with the true image minus the selectivity with the swapped image, with a paired image bootstrap.
The 500 base captions hold 3,711 object mentions, 3,177 grounded and 534 hallucinated.
All mentions align with their text position and label, and the label counts equal those of the CHAIR scorer for 500 of 500 captions.

\Cref{tab:tf} lists the result.
Only RLAIF-V separates grounded from hallucinated objects by a large margin, and only its separation depends on the image.
It also lowers the EOS log-probability at sentence ends by 9.5 nat, so it writes longer than the other models.
CHAIR-DPO, CSR and HA-DPO have a selectivity below 0.6 and no positive image dependence.

%% file: sec/S2_appendix_G_L.tex
\section{POPE and MMStar}
\label{app:G}

\begin{table*}[t]
  \centering
  \small
  \setlength{\tabcolsep}{5pt}
  \begin{tabular}{l l cc cc cc c}
    \toprule
    & & \multicolumn{2}{c}{POPE accuracy} & \multicolumn{2}{c}{Yes ratio (\%)} & \multicolumn{2}{c}{PO vs.\ yes-matched base} & MMStar \\
    \cmidrule(lr){3-4} \cmidrule(lr){5-6} \cmidrule(lr){7-8}
    Model & Split & Base & PO & Base & PO & $\Delta$Acc [95\% CI] & Agree (\%) & $\Delta$Acc [95\% CI] \\
    \midrule
    RLAIF-V   & r & 88.4 & 82.8 & 40.6 & 33.5 & $-$0.12 [$-$0.53, 0.24] & 96.9 & \multirow{3}{*}{0.73 [$-$1.20, 2.73]} \\
              & p & 87.2 & 82.3 & 41.8 & 34.0 & $-$0.20 [$-$0.59, 0.25] & 96.8 &  \\
              & a & 85.1 & 81.7 & 43.9 & 34.6 & $-$0.17 [$-$0.65, 0.42] & 96.5 &  \\
    \midrule
    CHAIR-DPO & r & 88.4 & 87.7 & 40.6 & 39.4 & 0.11 [$-$0.21, 0.53] & 98.8 & \multirow{3}{*}{$-$0.20 [$-$1.07, 0.67]} \\
              & p & 87.2 & 86.8 & 41.8 & 40.3 & 0.34 [0.01, 0.71] & 98.6 &  \\
              & a & 85.1 & 84.8 & 43.9 & 42.3 & 0.08 [$-$0.36, 0.57] & 98.2 &  \\
    \midrule
    CSR       & r & 88.4 & 88.4 & 40.6 & 40.4 & 0.16 [$-$0.04, 0.41] & 99.4 & \multirow{3}{*}{0.07 [$-$0.67, 0.87]} \\
              & p & 87.2 & 87.3 & 41.8 & 41.5 & 0.26 [0.03, 0.53] & 99.4 &  \\
              & a & 85.1 & 85.1 & 43.9 & 43.7 & 0.02 [$-$0.24, 0.35] & 99.2 &  \\
    \midrule
    HA-DPO    & r & 88.4 & 86.6 & 40.6 & 38.2 & $-$0.04 [$-$0.61, 0.47] & 95.3 & \multirow{3}{*}{0.00 [$-$1.40, 1.47]} \\
              & p & 87.2 & 85.7 & 41.8 & 39.2 & $-$0.14 [$-$0.68, 0.46] & 94.9 &  \\
              & a & 85.1 & 83.7 & 43.9 & 41.2 & $-$0.79 [$-$1.54, $-$0.01] & 94.1 &  \\
    \bottomrule
  \end{tabular}
  \caption{POPE and MMStar. Accuracies are given for the random (r), popular (p) and adversarial (a) POPE splits. $\Delta$Acc is the PO accuracy minus the yes-matched base accuracy with an image bootstrap interval, and Agree is the share of questions with the same answer. MMStar $\Delta$Acc is the PO accuracy minus the base accuracy (32.9) with a question bootstrap interval.}
  \label{tab:pope}
\end{table*}

On yes/no questions a model that says less answers yes less often.
We add one constant to the yes versus no log-odds of the base model so that its yes ratio matches the yes ratio of the PO model.
The constant is chosen on half of the images and applied to the other half with the cross-fitting of the main text, and both models answer by forced choice between yes and no.
\Cref{tab:pope} reports the accuracy difference and the share of questions with the same answer.
Over all four PO models and all three splits, the PO accuracy is within 0.79 points of the yes-matched base model, and the two models give the same answer on 94.1 to 99.4 percent of the questions.
Most of the POPE change of a PO model thus follows from its yes ratio.
On MMStar no PO model differs significantly from the base model.

\section{Paired differences between PO models}
\label{app:H}

\begin{table*}[t]
  \centering
  \small
  \setlength{\tabcolsep}{6pt}
  \begin{tabular}{l cc cc}
    \toprule
    & \multicolumn{2}{c}{CHAIR-500} & \multicolumn{2}{c}{AMBER} \\
    \cmidrule(lr){2-3} \cmidrule(lr){4-5}
    Pair & $\Draw$ & $\DS$ & $\Draw$ & $\DS$ \\
    \midrule
    RLAIF-V $-$ CHAIR-DPO & 0.51 [$-$0.58, 1.60] & 1.87 [0.27, 3.36] & 0.08 [$-$0.48, 0.61] & 2.02 [0.99, 2.92] \\
    RLAIF-V $-$ CSR & 2.09 [0.93, 3.28] & 2.98 [1.37, 4.61] & 1.37 [0.72, 2.06] & 3.68 [2.60, 4.70] \\
    CHAIR-DPO $-$ CSR & 1.59 [0.68, 2.52] & 1.11 [0.02, 2.34] & 1.29 [0.75, 1.82] & 1.66 [1.07, 2.32] \\
    RLAIF-V $-$ HA-DPO & 8.68 [6.13, 12.11] & 5.75 [2.86, 8.97] & 3.57 [2.82, 4.32] & 3.72 [2.70, 4.82] \\
    CHAIR-DPO $-$ HA-DPO & 8.17 [5.64, 11.47] & 3.88 [1.07, 7.26] & 3.50 [2.89, 4.11] & 1.70 [0.87, 2.66] \\
    CSR $-$ HA-DPO & 6.59 [4.06, 9.98] & 2.77 [$-$0.15, 5.98] & 2.21 [1.46, 2.93] & 0.04 [$-$0.84, 1.04] \\
    \bottomrule
  \end{tabular}
  \caption{Paired differences of the raw gain and of the stopping-controlled gain between PO models on the development sets, with 95 percent intervals from the same image bootstrap.}
  \label{tab:paired}
\end{table*}

\Cref{tab:paired} compares the PO models with each other on the same bootstrap draws.
On both benchmarks RLAIF-V keeps a larger stopping-controlled gain than CHAIR-DPO and CSR, although its raw gain is not significantly larger than that of CHAIR-DPO.
The ranking of PO models by raw gain and by stopping-controlled gain can therefore differ.

\section{Negative EOS probe}
\label{app:I}

\begin{figure*}[t]
  \centering
  \includegraphics[width=0.72\linewidth]{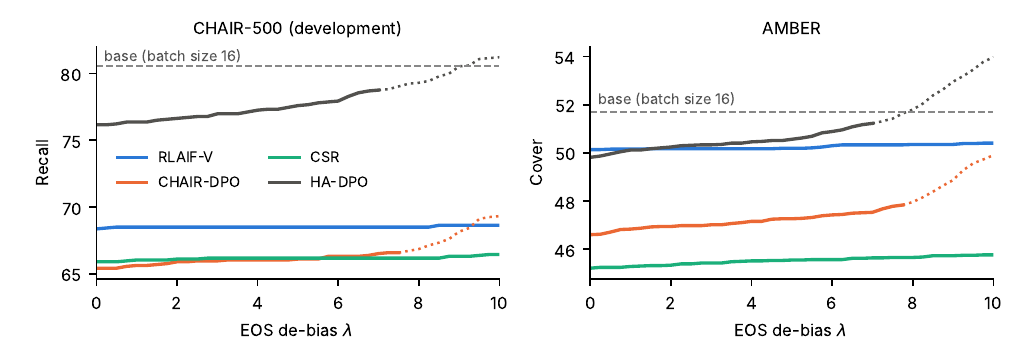}
  \caption{Recall (CHAIR-500) and Cover (AMBER) of the PO models along the negative EOS bias $\lambda$. Dotted parts have more than 5 percent of the captions at the 512-token limit. The dashed line is the base model at batch size 16.}
  \label{fig:debias}
\end{figure*}

\begin{table*}[t]
  \centering
  \footnotesize
  \setlength{\tabcolsep}{4pt}
  \begin{tabular}{l c c c c c c}
    \toprule
    Model & Recall drop & Largest recall ($\lambda$) & Recovered CHAIR / AMBER & Categories per caption & New categories & Words \\
    \midrule
    RLAIF-V   & 12.3 & 68.6 (10)   & 2\% / 17\% & 2.16 $\rightarrow$ 2.17 & 6 of 93 (6\% [2, 11])       & 93 $\rightarrow$ 98 \\
    CHAIR-DPO & 15.2 & 66.7 (7.65) & 9\% / 24\% & 2.06 $\rightarrow$ 2.13 & 38 of 989 (4\% [3, 5])      & 78 $\rightarrow$ 117 \\
    CSR       & 14.7 & 66.4 (10)   & 4\% / 8\%  & 2.17 $\rightarrow$ 2.21 & 19 of 529 (4\% [2, 5])      & 75 $\rightarrow$ 90 \\
    \midrule
    HA-DPO    & 4.5  & 78.8 (7.0)  & 58\% / 76\% & 2.83 $\rightarrow$ 3.04 & 101 of 1,224 (8\% [6, 11]) & 84 $\rightarrow$ 123 \\
    \bottomrule
  \end{tabular}
  \caption{Writing longer with a negative EOS bias on CHAIR-500, at the largest recall while less than 5 percent of the captions reach the 512-token limit. New categories counts the added object mentions that name a category not yet in the caption. The AMBER Cover drops are 1.6, 5.1, 6.5 and 1.9 points.}
  \label{tab:debias_full}
\end{table*}

\begin{table*}[t]
  \centering
  \footnotesize
  \setlength{\tabcolsep}{3pt}
  \begin{tabular}{l cccc cccc}
    \toprule
    & \multicolumn{4}{c}{CHAIR-500 recall} & \multicolumn{4}{c}{AMBER Cover} \\
    \cmidrule(lr){2-5} \cmidrule(lr){6-9}
    Model & Unreach. & Pooled gap [95\% CI] & Fold range & Limit (\%) & Unreach. & Pooled gap [95\% CI] & Fold range & Limit (\%) \\
    \midrule
    RLAIF-V   & 10 & $-$11.99 [$-$13.95, $-$9.88] & $-$13.56 to $-$10.35 & 0  & 7  & $-$1.34 [$-$2.37, $-$0.42] & $-$2.11 to $-$0.57 & 0 \\
    CHAIR-DPO & 10 & $-$11.30 [$-$13.12, $-$9.31] & $-$12.70 to $-$9.80  & 85 & 9  & $-$1.84 [$-$2.67, $-$1.09] & $-$2.85 to $-$0.82 & 83 \\
    CSR       & 10 & $-$14.18 [$-$16.09, $-$12.21] & $-$15.94 to $-$12.40 & 0 & 10 & $-$5.98 [$-$6.80, $-$5.16] & $-$6.77 to $-$5.20 & 0 \\
    HA-DPO    & 0  & $-$0.37 [$-$1.20, 0.29] & $-$2.82 to 1.72 & 61 & 0 & 0.04 [$-$0.12, 0.20] & $-$1.22 to 1.08 & 17 \\
    \bottomrule
  \end{tabular}
  \caption{Fold-level result of the negative EOS probe against the base run at batch size 16. Unreach.\ counts the calibration folds out of 10 with no $\lambda \leq 10$ within one point of the base value. The pooled gap is the PO value at the selected $\lambda$ minus the base value on the evaluation halves, Fold range gives the gaps of the 10 evaluation halves, and Limit is the share of captions at the 512-token limit at the selected $\lambda$.}
  \label{tab:debias_folds}
\end{table*}

\paragraph{Protocol.}
We decode each PO model once with an EOS bias of $-10$ and at most 512 new tokens at batch size 16, and we obtain every $\lambda \in [0, 10]$ with the post-hoc cut of \cref{app:B}.
The point $\lambda = 0$ comes from the same run, so the whole path shares one batch layout.
On each calibration half we choose the $\lambda$ whose recall is closest to the recall of a base run at the same batch size, and on the evaluation half we compare the two models.

\paragraph{Rules fixed before the run.}
We fixed five rules for RLAIF-V and CHAIR-DPO, with CHAIR-500 as the primary benchmark and AMBER as an independent check.
De-biasing would count as an intervention only if all five held.
\begin{enumerate}[leftmargin=*,itemsep=0pt,topsep=2pt]
  \item The pooled recall gap on the evaluation halves is within one point.
  \item Every calibration fold has a $\lambda \leq 10$ whose recall is within one point of the base recall. If no such $\lambda$ exists, we mark the fold as unreachable.
  \item Less than 5 percent of the captions reach the 512-token limit.
  \item The interval of the CHAIRi reduction lies above zero.
  \item The interval of the $\Hu$ reduction lies above zero.
\end{enumerate}

\paragraph{Outcome.}
RLAIF-V fails rules 1 and 2, and CHAIR-DPO fails rules 1, 2 and 3.
Their pooled recall stays 12.0 and 11.3 points below the base recall, and the target recall is unreachable in all ten CHAIR calibration folds for RLAIF-V, CHAIR-DPO and CSR.
On AMBER, 7 to 10 of the 10 folds are unreachable for these three models.
De-biasing is therefore not an intervention for them, and the main text uses the result as a diagnosis.
\Cref{fig:debias} shows the paths, and \cref{tab:debias_full} gives the point with the largest recall while less than 5 percent of the captions reach the limit. \Cref{tab:debias_folds} instead reports the $\lambda$ chosen to match the base value in each fold, where CHAIR-DPO reaches the limit for 85 percent of its captions.
The repeats are counted per COCO category, so a repeated category may refer to a new instance of the same object.
For example, at $\lambda = 10$ RLAIF-V adds ``Overall, the image captures a lovely moment between a bride and groom as they enjoy their cake together at a party.'' to a caption that already names the cake, the person and the tie.

\paragraph{Base coverage of the missed categories.}
On the CHAIR development set, the base caption names 47, 48 and 47 percent of the ground-truth categories that RLAIF-V, CHAIR-DPO and CSR miss, while the PO captions at $\lambda = 10$ name 2, 13 and 2 percent.
On AMBER the base caption covers 18, 15 and 16 percent of the missed objects, and the PO captions at $\lambda = 10$ cover 2, 6 and 1 percent.

\begin{table*}[!t]
  \centering
  \footnotesize
  \setlength{\tabcolsep}{4pt}
  \begin{tabular}{ll c cccc}
    \toprule
    Set & Model & Pairs & Tokens 0 to 20 & Tokens 20 to 60 & Tokens 60 to 100 & Later minus earlier \\
    \midrule
    Dev. & RLAIF-V & 369 & $-$0.25 [$-$0.31, $-$0.20] & 0.03 [$-$0.07, 0.12] & $-$0.16 [$-$0.23, $-$0.09] & $-$0.19 [$-$0.32, $-$0.04] \\
     & CHAIR-DPO & 248 & 0.00 [$-$0.04, 0.04] & $-$0.27 [$-$0.35, $-$0.19] & $-$0.25 [$-$0.34, $-$0.16] & 0.02 [$-$0.10, 0.14] \\
     & CSR & 229 & 0.02 [$-$0.04, 0.08] & $-$0.19 [$-$0.30, $-$0.09] & $-$0.28 [$-$0.37, $-$0.18] & $-$0.08 [$-$0.24, 0.07] \\
    \midrule
    Held-out & RLAIF-V & 374 & $-$0.35 [$-$0.41, $-$0.28] & 0.05 [$-$0.05, 0.13] & $-$0.15 [$-$0.23, $-$0.07] & $-$0.19 [$-$0.33, $-$0.06] \\
     & CHAIR-DPO & 256 & $-$0.04 [$-$0.11, 0.01] & $-$0.32 [$-$0.41, $-$0.22] & $-$0.19 [$-$0.30, $-$0.08] & 0.13 [$-$0.04, 0.30] \\
     & CSR & 233 & 0.00 [$-$0.06, 0.06] & $-$0.29 [$-$0.39, $-$0.19] & $-$0.23 [$-$0.34, $-$0.12] & 0.06 [$-$0.13, 0.24] \\
    \bottomrule
  \end{tabular}
  \caption{New ground-truth categories per caption in each token window, PO minus base, on the images where both captions reach 100 tokens. A category belongs to the window that holds the token piece that the scorer maps to its first mention. Later minus earlier is the difference of the two windows from 20 to 100 tokens. Intervals come from 1,000 image bootstrap draws.}
  \label{tab:discovery}
\end{table*}

\paragraph{Category discovery along the caption.}
We count the new ground-truth categories in two token windows, from 20 to 60 and from 60 to 100 tokens, on the images where both the base and the PO caption reach 100 tokens.
These are 46 to 75 percent of the images.
A category belongs to the window that holds the token piece that the scorer maps to its first mention, and we fixed the windows from the caption lengths of the development set before counting any category.
\Cref{tab:discovery} gives the paired difference between the PO and the base caption.
CHAIR-DPO and CSR name 0.19 to 0.32 fewer new ground-truth categories than the base model in both windows, and we detect no significant growth of this deficit in the later window.
In the first 20 tokens we detect no significant difference for these two models.
RLAIF-V names fewer categories in the first 20 tokens and in the later window, and the difference between its windows is below zero on both sets.
In the same windows, CHAIR-DPO and CSR keep 85 to 99 percent of the object mentions of the base model and name 39 to 68 percent of its new categories, counting hallucinated ones, while RLAIF-V mentions 61 to 63 percent of the base objects in the later window.
With the negative bias of \cref{tab:debias_full} on the development set, the ground-truth categories per caption grow by only 0.01 to 0.04 and stay 0.35 to 0.41 below the base run at batch size 16.

\paragraph{HA-DPO.}
HA-DPO loses only 4.5 recall points and reaches the base recall, but at the matched $\lambda$ (median 9.25) 60.6 percent of its CHAIR captions reach the length limit, so it fails rule 3.
At that point its CHAIRi reduction against the base model is 1.71 [0.04, 3.62], but its $\Hu$ reduction is $-1.18$ [$-3.45$, 2.02] and its CHAIRs reduction is $-2.0$ [$-6.3$, 4.3]. The CHAIRi reduction comes from repeated mentions that enlarge its denominator, since 83 percent of its object mentions at that point repeat a category of the same caption, against 60 percent at $\lambda = 0$.
On AMBER, 17 percent of its captions reach the limit and the remaining gain is $-0.04$ [$-1.26$, 0.83].

\section{Batch size and implementation}
\label{app:J}

Batched fp16 greedy decoding can flip near-tie tokens when the batch size changes, so every run uses one batch size from start to end, and a resumed run must have the same batch size, batch layout and input order.
On the development sets, the base model, RLAIF-V, CHAIR-DPO and CSR were decoded at batch size 32, and HA-DPO and the negative EOS runs at batch size 16. All held-out runs use batch size 16.
The development runs used A100 GPUs with 40 GB and 80 GB, and the held-out runs used one RTX PRO 6000 GPU. On 16 images, the batch size 16 captions of the two A100 types are identical.
\Cref{tab:bs} compares the base model at the two batch sizes.
The captions are identical for 435 of 500 CHAIR images and 889 of 1,004 AMBER images, and every reported gain moves by at most 0.15 points.
With the base run at batch size 16, the development interval of CSR on CHAIR moves from [$-0.03$, 2.75] to [0.16, 2.82].
Applied to development captions of all models at batch size 16, with the PO captions taken from the $\lambda = 0$ cut of the negative EOS runs, the frozen restoration thresholds of \cref{app:K} keep the recall within 0.27 points of the base recall and give $\Hu$ reductions of 5.19 [3.78, 6.82], 1.64 [0.64, 2.56], 1.07 [$-0.07$, 2.20] and 2.38 [0.68, 4.13] for RLAIF-V, CHAIR-DPO, CSR and HA-DPO.

\begin{table}[!t]
  \centering
  \footnotesize
  \setlength{\tabcolsep}{3pt}
  \begin{tabular}{l cc}
    \toprule
    Base model & Batch 32 & Batch 16 \\
    \midrule
    CHAIR CHAIRi / Recall & 14.39 / 80.5 & 14.42 / 80.6 \\
    CHAIR F1 / CHAIRs & 76.9 / 47.2 & 76.9 / 48.4 \\
    AMBER CHAIR / Cover & 7.50 / 51.8 & 7.53 / 51.7 \\
    \midrule
    $\DS$ RLAIF-V & 4.25 [2.62, 6.07] & 4.39 [2.66, 6.04] \\
    $\DS$ CHAIR-DPO & 2.39 [0.97, 4.13] & 2.54 [1.22, 4.19] \\
    $\DS$ CSR & 1.28 [$-$0.03, 2.75] & 1.42 [0.16, 2.82] \\
    $\DS$ HA-DPO & $-$1.49 [$-$4.40, 1.30] & $-$1.45 [$-$4.47, 1.36] \\
    \bottomrule
  \end{tabular}
  \caption{The base model decoded at batch size 32 and 16, and the stopping-controlled gains on CHAIR-500 against each base run. F1 is the set-level F1.}
  \label{tab:bs}
\end{table}

\section{Base-guided object-category restoration}
\label{app:K}

\begin{table*}[tp]
  \centering
  \footnotesize
  \setlength{\tabcolsep}{4pt}
  \begin{tabular}{l cccc}
    \toprule
    & RLAIF-V & CHAIR-DPO & CSR & HA-DPO \\
    \midrule
    Restoration recall / $\Hu$ / precision & 79.16 / 21.64 / 44.8 & 79.55 / 25.03 / 41.4 & 79.74 / 25.79 / 44.6 & 79.49 / 23.94 / 40.0 \\
    \midrule
    Random recall / $\Hu$ & 79.43 / 25.41 & 79.83 / 26.62 & 80.10 / 27.38 & 80.24 / 27.03 \\
    Restoration minus random, recall & $-$0.26 [$-$1.12, 0.58] & $-$0.28 [$-$0.90, 0.29] & $-$0.35 [$-$0.99, 0.22] & $-$0.76 [$-$1.44, $-$0.08] \\
    Random minus restoration, $\Hu$ & 3.77 [2.66, 4.92] & 1.59 [0.89, 2.37] & 1.59 [0.87, 2.31] & 3.09 [2.24, 3.93] \\
    Per-seed range & 3.13 to 4.09 & 1.23 to 1.91 & 0.92 to 2.09 & 2.55 to 3.81 \\
    \midrule
    Count-matched random recall & 77.27 & 78.84 & 78.78 & 78.69 \\
    Count-matched random minus restoration, $\Hu$ & 1.85 [0.78, 2.96] & 0.68 [$-$0.06, 1.48] & 0.91 [0.18, 1.67] & 0.75 [$-$0.04, 1.49] \\
    Test-calibrated random minus restoration, $\Hu$ & 3.47 [2.41, 4.62] & 1.22 [0.50, 1.99] & 1.58 [0.84, 2.33] & 1.73 [0.88, 2.49] \\
    \midrule
    Union recall / $\Hu$ / CHAIRs & 82.64 / 27.93 / 57.2 & 80.96 / 27.48 / 56.2 & 81.35 / 28.37 / 58.4 & 82.96 / 30.08 / 59.8 \\
    Union minus base, $\Hu$ & 1.05 [0.50, 1.60] & 0.60 [0.23, 1.00] & 1.49 [0.96, 2.03] & 3.20 [2.42, 4.02] \\
    Union minus restoration, $\Hu$ & 6.29 [5.16, 7.50] & 2.45 [1.72, 3.25] & 2.58 [1.85, 3.35] & 6.14 [5.08, 7.19] \\
    Restoration minus union, set-level F1 & 1.77 [1.02, 2.56] & 0.68 [0.11, 1.21] & 0.69 [0.13, 1.25] & 1.85 [1.03, 2.64] \\
    \midrule
    Category permutation recall / $\Hu$ / precision & 68.47 / 32.21 / 7.4 & 67.17 / 36.68 / 7.2 & 67.20 / 37.43 / 7.4 & 76.66 / 26.63 / 4.8 \\
    Category permutation minus restoration, $\Hu$ & 10.57 [9.03, 12.03] & 11.65 [10.23, 13.20] & 11.64 [10.25, 13.03] & 2.69 [1.92, 3.41] \\
    Co-occurrence prior recall / $\Hu$ / precision & 71.96 / 28.77 / 19.6 & 71.38 / 32.81 / 18.8 & 71.45 / 33.59 / 19.9 & 77.49 / 25.85 / 15.2 \\
    Co-occurrence prior minus restoration, $\Hu$ & 7.13 [5.53, 8.71] & 7.78 [6.32, 9.21] & 7.80 [6.34, 9.10] & 1.91 [1.18, 2.66] \\
    Shuffled-image base recall / $\Hu$ / precision & 68.03 / 32.67 / 5.8 & 66.54 / 37.37 / 5.4 & 66.64 / 38.06 / 5.7 & 76.64 / 26.66 / 4.6 \\
    Shuffled-image base minus restoration, $\Hu$ & 11.02 [9.46, 12.50] & 12.34 [10.85, 13.90] & 12.27 [10.88, 13.69] & 2.72 [1.95, 3.48] \\
    Frequency / text prior precision & 11.7 / 15.3 & 11.9 / 16.1 & 13.3 / 17.3 & 5.6 / 14.4 \\
    \bottomrule
  \end{tabular}
  \caption{Controls of restoration on the held-out set (base recall 80.13, base $\Hu$ 26.88). Except for the count-matched random rows, all controls were added after the frozen evaluation. Precision is the share of appended categories that are correct. The candidate-source rows use the number of appended categories of each image as the target, and the shuffled and permutation rows average 10 seeds. Intervals are 95 percent image bootstrap intervals. All entries use the pinned nltk 3.9.1.}
  \label{tab:controls}
\end{table*}

\paragraph{Procedure.}
For CHAIR the candidates are COCO categories found by the CHAIR extraction, and for AMBER they are noun lemmas of the AMBER hallucination vocabulary.
A candidate is a category of the base caption that the PO caption does not name.
Its score is the log-probability of the first token piece of its first mention, taken from the greedy pass of the base model.
Every candidate with a score strictly above $\tau$ is appended as ``There is also a \{object\} in the image.'' in score order, and the restored caption is scored again as text.
On the development set, $\tau$ is chosen by cross-fitting so that the restored captions reach the base recall, and a second threshold aims at the midpoint between the base and PO recalls.
For the held-out set we chose one threshold per model on the whole CHAIR development set and fixed it before generating the held-out images.
These thresholds are $\tau_0 = -1.290647$, $-1.707188$, $-1.616596$ and $-0.847897$ for RLAIF-V, CHAIR-DPO, CSR and HA-DPO, and the midpoint thresholds are $-0.496749$, $-0.581227$, $-0.568303$ and $-0.446861$.
No ground truth of the held-out set enters the selection. It is used only to score the outputs and to refit the stopping curve.
The CHAIR scorer maps the singular ``bus'' to ``bu'', as the original CHAIR code does, so an appended sentence for a bus is not counted.
This affects two appended categories each for CHAIR-DPO and CSR on the held-out set, and the counts below use the categories that the scorer finds in the restored caption.

\paragraph{Base-only filtering on the development sets.}
Base-only filtering keeps the base caption and removes its objects with a score below a threshold.
It cannot exceed the base recall.
At the midpoint recall, its $\Hu$ is above the base stopping curve by 4.19, 5.27, 4.94 and 1.15 points on CHAIR and by 0.28, 1.68, 2.11 and 0.42 points on AMBER for RLAIF-V, CHAIR-DPO, CSR and HA-DPO.
The base model alone thus does not reach the trade-off of restoration, which starts from the precise PO caption.

\paragraph{Held-out controls.}
\Cref{tab:controls} lists the controls on the held-out set.
Random restoration appends each base-only candidate with a probability $q$ fixed on the development set so that the seed-averaged development recall reaches the base recall, with $q = 0.805$, 0.925, 0.915 and 0.605. We apply it with 10 new seeds.
Its intervals come from the image bootstrap of the seed-averaged counts, and we report the per-seed range separately.
Two further rows serve only as diagnostics. Count-matched random restoration admits random candidates until their total number over the set equals that of restoration, and test-calibrated random restoration tunes the number of candidates on the held-out ground truth to the restored recall. Both average 10 seeds.

\input{sec/S4_table_sweep}

\paragraph{Effect on CHAIRi and accuracy of the appended categories.}
Against the base model, restoration lowers CHAIRi on the held-out set by 4.79, 3.45 and 2.53 points for RLAIF-V, CHAIR-DPO and CSR, and the number of distinct hallucinated objects per image by 0.24, 0.09 and 0.05.
The PO captions of these models miss 523, 551 and 549 ground-truth categories, and the base caption names 48, 46 and 47 percent of them.
Only 38.5, 39.3 and 41.8 percent of all base-only candidates are such recoverable categories, while 73.1 percent of all categories in the base captions are correct.
The correct appended categories are 79, 91 and 90 percent of these recoverable categories, and \cref{tab:sweep} gives the number and precision of the appended categories at $\delta = 0$.

\paragraph{Accuracy of the confidence score.}
Internal representations of LVLMs also give a higher confidence to real objects than to hallucinated ones~\cite{jiang2025projectaway}.
On the CHAIR development set, the base confidence separates correct from hallucinated base-only candidates with an AUC of 0.68 to 0.69.
With a swapped image, the same base log-probabilities are lower by 1.6 to 1.9 nat on average and reach an AUC of only 0.52 to 0.54, so the base confidence uses the image.
The signed difference of the true-image minus the swapped-image log-probability is a weak predictor of the label, with an AUC of 0.52 to 0.53 for the base model and 0.52 to 0.57 for the PO models.

\begin{figure}[!t]
  \centering
  \footnotesize
  \newcommand{\ok}[1]{#1\,\checkmark}
  \newcommand{\no}[1]{#1\,$\times$}
  \includegraphics[width=0.6\linewidth]{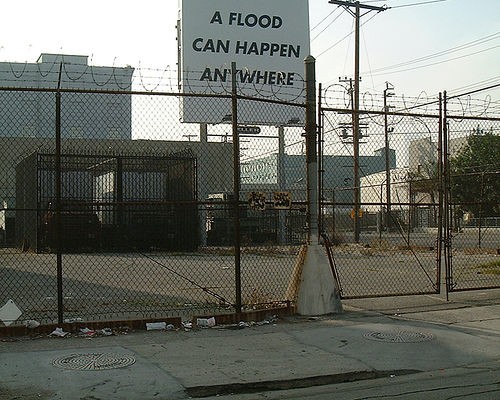}\par\vspace{3pt}
  \setlength{\tabcolsep}{2pt}
  \begin{tabular}{@{}l >{\raggedright\arraybackslash}p{0.72\linewidth}@{}}
    GT & truck \\
    Base (99 w.) & \no{car}, \ok{truck}, \no{fire hydrant}, \no{traffic light} \\
    PO (82 w.) & no COCO category \\
    PO, $\lambda{=}10$ & no COCO category (94 w.) \\
    Restored & $+$ \ok{truck} ($-0.83$), \no{car} ($-1.18$), \no{fire hydrant} ($-1.23$) \\
    Below $\tau$ & \no{traffic light} ($-1.36$), $\tau = -1.29$ \\
  \end{tabular}
  \caption{A failure case of restoration for RLAIF-V on CHAIR-500 (COCO 300032), in the format of \cref{fig:qual}.}
  \label{fig:fail}
\end{figure}

\paragraph{A failure case.}
\Cref{fig:fail} shows a development image where restoration appends hallucinated categories.
The base model is confident about the cars and the fire hydrant of a parking-lot scene that it imagines, so restoration appends two hallucinated categories together with the missed truck.

\input{sec/S3_table_recal}

\begin{figure*}[tp]
  \centering
  \includegraphics[width=0.95\linewidth]{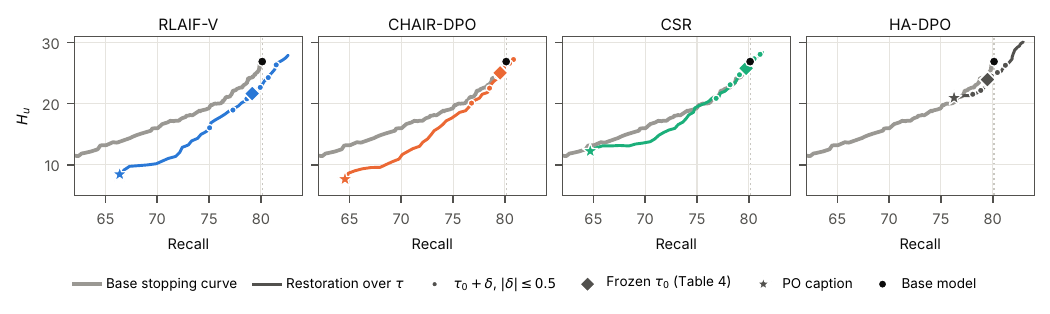}
  \caption{Distinct-object hallucination rate against recall on the held-out set. The gray curve is the base stopping curve, the colored line is restoration over a dense sweep of $\tau$, dots mark $\tau_0 + \delta$ for the $\delta$ of \cref{tab:sweep}, the diamond marks $\tau_0$, the star marks the PO model and the black dot marks the base model.}
  \label{fig:sweep}
\end{figure*}

\paragraph{Matched-recall comparison and threshold sweep.}
\Cref{tab:recal} compares restoration with the base stopping curve refitted on the held-out set, and \cref{tab:sweep,fig:sweep} sweep the threshold over $\tau_0 \pm 0.5$.
Both analyses were added after the frozen evaluation and do not change the acceptance decision of the main text.
The selected EOS bias lies on a grid with a step of 0.05, so the curve recall differs from the restored recall by at most 0.17 points.

\paragraph{Candidate-source controls.}
The lower rows of \cref{tab:controls} replace the source of the appended categories and use the number of categories that the frozen restoration appends to each image as the target.
The shuffled-image base takes the base-only candidates of other held-out images in their own confidence order.
The category permutation deals the categories that restoration appended over the whole set to the images again at random.
The frequency prior ranks categories by how often they are missed ground truth in the PO captions of the development set, and the co-occurrence and text priors condition this rate on the categories or the content words of the PO caption of the same image, with a shrinkage weight chosen by cross-fitting on the development set.
No control appends a category that the PO caption already names, and the ground truth of the held-out images is used only for scoring.
Every control has a lower precision and a higher $\Hu$ than restoration.
The intervals of the $\Hu$ differences lie above zero, and the recall differences have interval lower bounds of 5.57 points or more for RLAIF-V, CHAIR-DPO and CSR and 1.23 points or more for HA-DPO.
The base-only candidates of the same image are thus more useful than these tested sources.
These controls are post-hoc diagnostics.
A first run after the frozen evaluation had the shuffled base and the frequency and co-occurrence priors, and we then added the text prior and the category permutation and widened the grid of the shrinkage weight.
They take the number of categories per image from the frozen restoration and cannot be deployed.
The scorer-added counts differ slightly from these targets, since the scorer misses the singular bus in two restored captions each of CHAIR-DPO and CSR and the permutation leaves 0 to 3 slots unfilled per seed.
The intervals of the shuffled base and the permutation are conditional on their 10 seeds, and the per-seed range of their $\Hu$ difference is 10.3 to 12.9 points for RLAIF-V, CHAIR-DPO and CSR.

\paragraph{Diagnosed misses and image-level outcome.}
\Cref{tab:diag} links restoration to the diagnosis of Sec.~\ref{sec:recall}.
On the development set we apply the frozen $\tau_0$ to the PO captions at $\lambda = 0$ of the negative EOS run at batch size 16, which is a post-hoc analysis.
Condition E marks a base-only ground-truth category that is still missing at the $\lambda$ of \cref{tab:debias_full}, which gives the largest recall while less than 5 percent of the captions reach the 512-token limit, with $\lambda = 10$, 7.65 and 10 for RLAIF-V, CHAIR-DPO and CSR.
Condition S marks a category that the recall-matched positive EOS stop of the base model also removes.
Of the base-only ground-truth categories, 92 to 98 percent meet condition E, and restoration fills 81 to 92 percent of them.
For the categories that meet both S and E, the restored share is 79 to 93 percent, and we detect no significant difference to the other base-only categories.
This share counts base-only categories. Over all ground-truth categories that the PO captions miss on the held-out set, restoration fills 38.0, 42.3 and 42.6 percent.
The appended categories that meet condition S have a lower precision than the others, and they account for 63 to 78 percent of the appended hallucinations.
On the held-out set, 30 to 36 percent of the images gain a ground-truth category, 23 to 26 percent gain only hallucinated categories and 39 to 46 percent gain no new category, mostly because the PO caption already names every base category.

\begin{table}[t]
  \centering
  \scriptsize
  \setlength{\tabcolsep}{2pt}
  \begin{tabular}{>{\raggedright\arraybackslash}p{2.7cm} ccc}
    \toprule
    & RLAIF-V & CHAIR-DPO & CSR \\
    \midrule
    \multicolumn{4}{l}{Development set, batch size 16, post hoc} \\
    Base-only GT / restored & 216 / 175 & 240 / 222 & 233 / 212 \\
    Share with E (\%) & 98 & 92 & 97 \\
    Restored with E (\%) & 81 [75, 87] & 92 [89, 96] & 91 [87, 95] \\
    Restored with S and E (\%) & 79 & 93 & 92 \\
    \quad minus other types (pt) & $-$3 [$-$16, 7] & 1 [$-$7, 8] & 2 [$-$5, 10] \\
    \midrule
    \multicolumn{4}{l}{Held-out set, frozen restoration} \\
    Precision, S / other (\%) & 33.1 / 57.4 & 34.5 / 54.3 & 38.7 / 58.7 \\
    Hallucinations with S (\%) & 63 [54, 70] & 73 [67, 79] & 78 [72, 84] \\
    Images, GT gain / only hallucinations / none (\%) & 30 / 23 / 46 & 35 / 26 / 39 & 36 / 24 / 40 \\
    \bottomrule
  \end{tabular}
  \caption{Diagnosed misses and restored categories. Base-only GT are ground-truth categories that the base caption names and the PO caption misses. E marks those still missing at the negative EOS bias of \cref{tab:debias_full} in the same run, the largest recall while less than 5 percent of the captions reach the 512-token limit. S marks those that the recall-matched positive EOS stop of the base model also removes, averaged over the five cross-fitting splits. Counts use the categories that the CHAIR scorer finds in the restored caption. Images with a ground-truth gain may also gain hallucinated categories. Intervals come from 1,000 image bootstrap draws.}
  \label{tab:diag}
\end{table}

\section{Held-out decomposition}
\label{app:L}

The held-out set has 500 COCO val2014 images drawn with seed 1 and no overlap with CHAIR-500.
All five models were decoded at batch size 16, and the EOS bias of the stopping curve was refitted on these images by cross-fitting.
\Cref{tab:heldout} repeats the decomposition of the main text.
The stopping share is 58, 61 and 74 percent for RLAIF-V, CHAIR-DPO and CSR, against 61, 77 and 86 percent on CHAIR-500. All three keep a stopping-controlled gain whose interval excludes zero.

\begin{table}[!tb]
  \centering
  \scriptsize
  \setlength{\tabcolsep}{1.9pt}
  \begin{tabular}{l ccc c c c}
    \toprule
    Model & CHAIRi & Recall & $\Hu$ & $\Draw$ & $\DS$ [95\% CI] & $s$ \\
    \midrule
    LLaVA-1.5-7B & 15.91 & 80.13 & 26.88 & & & \\
    RLAIF-V   & 4.05 & 66.37 & 8.43  & 11.86 & 4.99 [3.53, 6.43] & 58\% \\
    CHAIR-DPO & 4.07 & 64.57 & 7.64  & 11.84 & 4.61 [3.10, 6.11] & 61\% \\
    CSR       & 6.29 & 64.69 & 12.22 & 9.62  & 2.46 [1.10, 3.58] & 74\% \\
    HA-DPO    & 11.74 & 76.27 & 20.93 & 4.18 & 1.09 [$-$0.39, 3.06] & 74\% \\
    \bottomrule
  \end{tabular}
  \caption{Decomposition of the CHAIRi gain on the held-out set.}
  \label{tab:heldout}
\end{table}

%% file: sec/S4_table_sweep.tex
\begin{table*}[tp]\centering\scriptsize\setlength{\tabcolsep}{3.6pt}
\begin{tabular}{l r cccc cc cc}\toprule
Model & $\delta$ & Recall & $\Hu$ & CHAIRi & F1 & Appended & Precision (\%) & $\Delta\Hu$ vs.\ base & Gap to curve (recalibrated) \\ \midrule
RLAIF-V & $-$0.50 & 81.48 & 26.34 & 14.00 & 77.4 & 594 & 39.6 & 0.54 [$-$0.30, 1.44] & recall above base \\
 & $-$0.25 & 80.71 & 24.26 & 12.72 & 78.1 & 531 & 42.0 & 2.62 [1.61, 3.63] & recall above base \\
 & $-$0.10 & 79.94 & 22.65 & 11.74 & 78.6 & 480 & 44.0 & 4.23 [3.06, 5.43] & 2.93 [0.68, 4.77] \\
 & 0.00 & 79.16 & 21.64 & 11.12 & 78.8 & 444 & 44.8 & 5.24 [3.92, 6.60] & 3.10 [0.31, 5.21] \\
 & +0.10 & 78.65 & 20.79 & 10.62 & 78.9 & 417 & 45.8 & 6.09 [4.79, 7.54] & 2.84 [0.30, 5.65] \\
 & +0.25 & 77.30 & 18.95 & 9.55 & 79.1 & 356 & 47.8 & 7.93 [6.50, 9.57] & 2.51 [0.22, 6.05] \\
 & +0.50 & 75.05 & 16.04 & 7.93 & 79.3 & 263 & 51.3 & 10.83 [9.08, 12.77] & 3.39 [0.96, 6.36] \\
\midrule
CHAIR-DPO & $-$0.50 & 80.84 & 27.21 & 13.73 & 76.6 & 642 & 39.6 & $-$0.34 [$-$0.82, 0.14] & recall above base \\
 & $-$0.25 & 80.45 & 26.58 & 13.36 & 76.8 & 619 & 40.1 & 0.29 [$-$0.33, 0.96] & recall above base \\
 & $-$0.10 & 79.87 & 25.85 & 12.92 & 76.9 & 590 & 40.5 & 1.03 [0.21, 1.82] & $-$0.27 [$-$2.02, 0.93] \\
 & 0.00 & 79.55 & 25.03 & 12.46 & 77.2 & 565 & 41.4 & 1.85 [0.98, 2.77] & 0.18 [$-$1.68, 1.58] \\
 & +0.10 & 79.29 & 24.54 & 12.19 & 77.3 & 549 & 41.9 & 2.34 [1.35, 3.31] & 0.36 [$-$1.65, 1.85] \\
 & +0.25 & 78.52 & 22.53 & 11.10 & 78.0 & 490 & 44.5 & 4.35 [3.27, 5.49] & 1.01 [$-$1.08, 3.12] \\
 & +0.50 & 76.78 & 20.08 & 9.79 & 78.3 & 408 & 46.8 & 6.80 [5.38, 8.19] & 0.93 [$-$1.14, 3.63] \\
\midrule
CSR & $-$0.50 & 81.09 & 28.07 & 14.72 & 76.2 & 609 & 42.0 & $-$1.19 [$-$1.80, $-$0.60] & recall above base \\
 & $-$0.25 & 80.51 & 27.34 & 14.27 & 76.4 & 579 & 42.7 & $-$0.46 [$-$1.19, 0.27] & recall above base \\
 & $-$0.10 & 80.00 & 26.26 & 13.65 & 76.7 & 543 & 44.0 & 0.62 [$-$0.24, 1.49] & $-$0.40 [$-$2.43, 0.72] \\
 & 0.00 & 79.74 & 25.79 & 13.38 & 76.9 & 527 & 44.6 & 1.08 [0.15, 2.05] & $-$0.28 [$-$2.38, 1.10] \\
 & +0.10 & 79.10 & 24.72 & 12.77 & 77.1 & 490 & 45.9 & 2.15 [1.07, 3.22] & $-$0.17 [$-$2.37, 1.53] \\
 & +0.25 & 78.20 & 23.09 & 11.86 & 77.6 & 436 & 48.4 & 3.79 [2.58, 5.11] & $-$0.11 [$-$2.55, 2.30] \\
 & +0.50 & 76.85 & 20.86 & 10.66 & 78.0 & 365 & 52.1 & 6.02 [4.53, 7.48] & 0.28 [$-$2.01, 3.07] \\
\midrule
HA-DPO & $-$0.50 & 81.22 & 26.27 & 14.66 & 77.3 & 213 & 36.1 & 0.61 [$-$0.63, 1.84] & recall above base \\
 & $-$0.25 & 80.45 & 25.09 & 14.00 & 77.6 & 170 & 38.2 & 1.79 [0.43, 3.10] & recall above base \\
 & $-$0.10 & 79.87 & 24.50 & 13.67 & 77.6 & 145 & 38.6 & 2.38 [0.91, 3.79] & 1.08 [$-$1.63, 2.84] \\
 & 0.00 & 79.49 & 23.94 & 13.36 & 77.7 & 125 & 40.0 & 2.94 [1.46, 4.47] & 1.22 [$-$1.68, 3.10] \\
 & +0.10 & 79.42 & 23.10 & 12.94 & 78.1 & 106 & 46.2 & 3.78 [2.25, 5.33] & 1.97 [$-$0.91, 3.92] \\
 & +0.25 & 78.84 & 22.16 & 12.45 & 78.3 & 75 & 53.3 & 4.72 [3.07, 6.29] & 1.80 [$-$0.95, 4.39] \\
 & +0.50 & 78.07 & 21.53 & 12.11 & 78.3 & 47 & 59.6 & 5.35 [3.63, 6.92] & 1.15 [$-$1.57, 4.12] \\
\bottomrule\end{tabular}
\caption{Sweep of the restoration threshold $\tau = \tau_0 + \delta$ on the held-out set. Row $\delta = 0$ is the frozen result of the main text. Appended is the number of appended categories and Precision is the share of them in the ground truth. $\Delta\Hu$ vs.\ base is the base value minus the restored value with an image bootstrap interval. The gap to the recall-matched stopping curve uses the recalibrated bootstrap and exists only while the restored recall is below the base recall of 80.13. The rows share the same images and are correlated.}
\label{tab:sweep}\end{table*}

%% file: sec/S3_table_recal.tex
\begin{table*}[tp]\centering\footnotesize\setlength{\tabcolsep}{4pt}
\begin{tabular}{ll cc cc ccc}\toprule
Model & Threshold & Restored recall & Curve recall & Restored $\Hu$ & Curve $\Hu$ & Gap & Conditional CI & Recalibrated CI \\ \midrule
RLAIF-V & $\tau_0$ & 79.16 & 79.20 & 21.64 & 24.75 & 3.10 & [1.73, 4.56] & [0.31, 5.21] \\
 & Midpoint & 72.28 & 72.31 & 12.12 & 17.54 & 5.42 & [3.38, 7.53] & [2.74, 8.05] \\
\midrule
CHAIR-DPO & $\tau_0$ & 79.55 & 79.50 & 25.03 & 25.21 & 0.18 & [$-$0.81, 1.24] & [$-$1.68, 1.58] \\
 & Midpoint & 70.55 & 70.42 & 12.03 & 16.39 & 4.36 & [2.44, 6.28] & [2.12, 6.87] \\
\midrule
CSR & $\tau_0$ & 79.74 & 79.58 & 25.79 & 25.51 & $-$0.28 & [$-$1.23, 0.75] & [$-$2.38, 1.10] \\
 & Midpoint & 71.32 & 71.15 & 14.89 & 16.80 & 1.91 & [0.12, 3.67] & [$-$0.25, 4.51] \\
\midrule
HA-DPO & $\tau_0$ & 79.49 & 79.45 & 23.94 & 25.15 & 1.22 & [$-$0.28, 2.67] & [$-$1.68, 3.10] \\
 & Midpoint & 78.33 & 78.21 & 21.67 & 23.21 & 1.54 & [$-$0.21, 3.18] & [$-$1.42, 4.25] \\
\bottomrule\end{tabular}
\caption{Restoration against the base stopping curve at matched recall on the held-out set. Gap is the curve $\Hu$ minus the restored $\Hu$, so a positive gap favors restoration. The curve is refitted on the held-out set with the cross-fitting of the main text. The conditional interval keeps the selected stopping policy fixed and resamples the images. The recalibrated interval repeats the choice of the EOS bias in every one of 1,000 bootstrap draws, and the main text reports it. The curve recall differs from the restored recall by the grid step of the EOS bias.}
\label{tab:recal}\end{table*}